\documentclass{article}

    \PassOptionsToPackage{numbers, compress}{natbib}
\usepackage[preprint]{neurips_2022}

\usepackage[utf8]{inputenc} % allow utf-8 input
\usepackage[T1]{fontenc}    % use 8-bit T1 fonts
\usepackage{hyperref}       % hyperlinks
\usepackage{url}            % simple URL typesetting
\usepackage{booktabs}       % professional-quality tables
\usepackage{amsfonts}       % blackboard math symbols
\usepackage{nicefrac}       % compact symbols for 1/2, etc.
\usepackage{microtype}      % microtypography
\usepackage{xcolor}         % colors
\usepackage{amsmath}
\usepackage{amssymb}
\usepackage{mathtools}
\usepackage{amsthm}
\usepackage{arydshln}
\usepackage{subfig}
\usepackage{wrapfig}
\usepackage{multirow}
\usepackage{tikz}
\usepackage{tikz-cd}
\usepackage{algorithmic}
\usepackage{algorithm}
\theoremstyle{plain}
\newtheorem{theorem}{Theorem}[section]

\theoremstyle{definition}

\theoremstyle{remark}

\newcommand{\norm}[1]{\left\lVert#1\right\rVert}
\def\multiset#1#2{\ensuremath{\left(\kern-.3em\left(\genfrac{}{}{0pt}{}{#1}{#2}\right)\kern-.3em\right)}}
\usepackage[textsize=tiny]{todonotes}
\usepackage{pifont} 
\newcommand{\cmark}{\ding{51}}
\newcommand{\xmark}{\ding{55}}

\title{A new perspective on building efficient and expressive 3D equivariant graph neural networks}

\author{
  Weitao Du$^1$\thanks{Equal contribution.} \\
  \And
  Yuanqi Du$^2$\footnotemark[1] \\
  \And
  Limei Wang$^3$\footnotemark[1] \\
  \And
  Dieqiao Feng$^2$ \\  
  \And 
  Guifeng Wang$^4$\\
  \AND 
  Shuiwang Ji$^3$\\
  \And 
  Carla P Gomes$^2$\\
  \And 
  Zhi-Ming Ma$^1$ \\
  \AND
  \vspace{-5mm}\\
  $^1$ Chinese Academy of Sciences \\
  $^2$ Cornell University \\
  $^3$ Texas A\&M University\\
  $^4$ Zhejiang University\\
}

\begin{document}

\maketitle

\begin{abstract}
Geometric deep learning enables the encoding of physical symmetries in modeling 3D objects. Despite rapid progress in encoding 3D symmetries into Graph Neural Networks (GNNs), a comprehensive evaluation of the expressiveness of these networks through a local-to-global analysis lacks today. In this paper, we propose a local hierarchy of 3D isomorphism to evaluate the expressive power of equivariant GNNs and investigate the process of representing global geometric information from local patches. Our work leads to two crucial modules for designing expressive and efficient geometric GNNs; namely local substructure encoding (\textbf{LSE}) and frame transition encoding (\textbf{FTE}). To demonstrate the applicability of our theory, we propose LEFTNet which effectively implements these modules and achieves state-of-the-art performance on both scalar-valued and vector-valued molecular property prediction tasks. We further point out the design space for future developments of equivariant graph neural networks. Our codes are available at \url{https://github.com/yuanqidu/LeftNet}.
\end{abstract}

\section{Introduction}
\label{sec:intro}

The success of many deep neural networks can be attributed to their ability to respect physical symmetry, such as Convolutional Neural Networks (CNNs)~\cite{he2016deep} and Graph Neural Networks (GNNs)~\cite{kipf2016semi}. Specifically, CNNs encode translation equivariance, which is essential for tasks such as object detection. 
% by the parameter-sharing kernels. 
Similarly, GNNs encode permutation equivariance, which ensures that the node ordering does not affect the output node representations.
, by aggregating neighboring messages. 
Modeling 3D objects, such as point clouds and molecules, is a fundamental problem with numerous applications, including robotics~\cite{simeonov2022neural}, molecular simulation~\cite{noe2020machine,holdijk2022path}, and drug discovery~\cite{bronstein2017geometric,atz2021geometric,wang2022advanced,schneuing2022structure,du2022molgensurvey}. Different from 2D pictures and graphs that only possess the translation~\cite{he2016deep} and permutation~\cite{kipf2016semi} symmetry, 3D objects intrinsically encode the complex $SE(3)/ E(3)$ symmetry~\cite{bronstein2021geometric}, which makes their modeling a nontrivial task in the machine learning community.

To tackle this challenge, several approaches have been proposed to effectively encode 3D rotation and translation equivariance in the deep neural network architectures, such as TFN~\cite{thomas2018tensor}, EGNN~\cite{satorras2021n}, and SphereNet~\cite{liu2021spherical}. TFN leverages spherical harmonics to represent and update tensors equivariantly, while EGNN processes geometric information through vector update. On the other hand, SphereNet is invariant by encoding scalars like distances and angles.
Despite rapid progress has been made on the empirical side, it's still unclear what 3D geometric information can equivariant graph neural networks capture and how the geometric information is integrated during the message passing process~\cite{gilmer2017neural, battaglia2018relational, yang2019analyzing}. 
This type of analysis is crucial in designing expressive and efficient 3D GNNs, as it's usually a trade-off between encoding enough geometric information and preserving relatively low computation complexity. Put aside the $SE(3)/ E(3)$ symmetry, this problem is also crucial in analysing ordinary GNNs. For example, 1-hop based message passing graph neural networks~\cite{xupowerful} are computationally efficient while suffering from expressiveness bottlenecks (comparing with subgraph GNNs~\cite{zhaostars,frasca2022understanding}). On the other hand, finding a better trade-off for 3D GNNs is more challenging, since we must ensure that the message updating and aggregating process respects the $SE(3)/ E(3)$ symmetry.

In this paper, we attempt to discover better trade-offs between computational efficiency and expressiveness power for 3D GNNs by studying two specific questions: 1. What is the expressive power of invariant scalars in encoding 3D geometric patterns? 2. Is equivariance really necessarily for 3D GNNs?  The first question relates to the design of node-wise geometric messages, and the second question relates to the design of equivariant (or invariant) aggregation. To tackle these two problems, we take a local-to-global approach. 
More precisely, we first define three types of 3D isomorphism to characterize local 3D structures: tree, triangular, and subgraph isomorphism, following a local hierarchy. As we will discuss in the related works section, our local hierarchy lies between the 1-hop and 2-hop geometric isomorphism defined in~\cite{joshi2022expressive}. Then, we can measure the expressiveness power of 3D GNNs by their ability of differentiating non-isomorphic 3D structures in a similar way as the geometric WL tests in~\cite{joshi2022expressive}.
Under this theoretical framework, we summarize one essential ingredient for building expressive geometric messages on each node: local 3D substructure encoding (\textbf{LSE}), which allows an invariant realization. To answer the second question, we analyze  whether local invariant features are sufficient for expressing global geometries by message aggregation, and it turns out that frame transition encoding (\textbf{FTE}) is crucial during the local to global process. Although \textbf{FTE} can be realized by invariant scalars, we further demonstrate that introducing equivariant messaging passing is more efficient.   By connecting \textbf{LSE} and \textbf{FTE} modules, we are able to present a modular overview of 3D GNNs designs.

In realization of our theoretical findings, we propose LEFTNet that efficiently implements \textbf{LSE} and \textbf{FTE} (with equivariant tensor update) without sacrificing expressiveness. Empirical experiments on real-world scenarios, predicting scalar-valued property (e.g. energy) and vector-valued property (e.g. force) for molecules, demonstrate the effectiveness of LEFTNet.

\section{Preliminary}
\label{sec:prelim}

In this section, we provide an overview of the mathematical foundations of $E(3)$ and $SE(3)$ symmetry, which is essential in modeling 3D data. We also summarize the message passing graph neural network framework, which enables the realization of $E(3)/SE(3)$ equivariant models. 

\textbf{Euclidean Symmetry.} Our target is to incorporate Euclidean symmetry to ordinary permutation-invariant graph neural networks. The formal way of describing Euclidean symmetry is the group $E(3) = O(3) \rtimes T(3)$, where $O(3)$ corresponds to reflections (parity transformations) and rotations. For tasks that are anti-symmetric under reflections (e.g. chirality), we consider the subgroup $SE(3)= SO(3) \rtimes T(3)$, where $SO(3)$ is the group of rotations. 
We will use $SE(3)$ in the rest of the paper for brevity except when it's necessary to emphasize reflections. 

\textbf{Equivariance.} A tensor-valued function $f(\textbf{x})$ is said to be \textbf{equivariant} with respect to $SE(3)$ if for any translation or rotation $g \in SE(3)$ acting on $\textbf{x} \in \mathbf{R}^3$, we have $$f(g \textbf{x}) = \mathcal{M}(g) f(\textbf{x}),$$
where $\mathcal{M}(\cdot)$ is a matrix representation of $SE(3)$ acting on tensors. See Appendix \ref{appendix:back} for a general definition of tensor fields. In this paper, we will use \textbf{bold} letters to represent an equivariant tensor, e.g., $\textbf{x}$ as a position vector. It is worth noting that when $f(\textbf{x}) \in \mathbf{R}^1$ and $\mathcal{M}(g) \equiv 1$ (the constant group representation), the equivariant function $f(\textbf{x})$ is also called an \textbf{invariant} scalar function. 

\textbf{Scalarization.} Scalarization is a general technique that originated from differential geometry for realizing covariant operations on tensors~\cite{hsu2002stochastic}. Our method will apply a simple version of scalarization in $\textbf{R}^3$ to transform equivariant quantities.  
At the heart of its realization is the notion of equivariant orthonormal frames, which consist of three orthonormal equivariant vectors:
$$\mathcal{F} : = (\textbf{e}_1,\textbf{e}_2,\textbf{e}_3).$$
Based on $\mathcal{F}$, we can build orthonormal equivariant frames for higher order tensors by taking tensor products $\otimes$, see Eq.~\ref{higher sca} in Appendix.
By taking the inner product between $\mathcal{F}$ and a given equivariant vector (tensor) $\textbf{x}$, we get a tuple of invariant scalars (see \cite{du2022se} for a proof):
\begin{equation} \label{scalarization}
\textbf{x} \rightarrow \Tilde{x}: = (\textbf{x}\cdot \textbf{e}_1,\textbf{x}\cdot \textbf{e}_2, \textbf{x}\cdot \textbf{e}_3),\end{equation} 
and $\Tilde{x}$ can be seen as the `scalarized' coordinates of $\textbf{x}$. 

\textbf{Tensorization.} Tensorization, on the other hand, is the `reverse' process of scalarization. Given a tuple of scalars: $(x_1,x_2,x_3)$, tensorization creates an equivariant vector (tensor) out of $\mathcal{F}$:
\begin{equation} \label{tensorization}
(x_1,x_2,x_3) \xrightarrow{\text{Pairing}}\textbf{x}:= x_1\textbf{e}_1 + x_2\textbf{e}_2 + x_3\textbf{e}_3.\end{equation}
The same procedure is extended to higher order cases, see Eq.~\ref{higher tensorization} in Appendix. 

\textbf{Message Passing Scheme for Geometric Graphs.} A geometric graph $G$ is represented by $G = (V,E)$. Here, $v_i \in V$ denotes the set of nodes (vertices, atoms), and $e_{ij} \in E$ denotes the set of edges. 
% Denote the average degree of each node by $k$. 
For brevity, the edge feature attached on $e_{ij}$ is also denoted by $e_{ij}$. 
Let $\textbf{X}=(\textbf{x}_1, \dots, \textbf{x}_n) \in \mathbf{R}^{n \times 3}$ be the 3D point cloud of all nodes' equivariant positions, which determines the 3D geometric structure of $G$.

A common machine learning tool for modeling graph-structured data is the Message Passing Neural Network (MPNN)~\cite{gilmer2017neural}. A typical 1-hop MPNN framework consists of two phases: (1) message passing; (2) readout. Let $h_i^l, h_j^l$ denote the $l$-th layer's node features of source $i$ and target $j$ that also depend on the 3D positions $(\textbf{x}_i,\textbf{x}_j)$, then the aggregated message is % has the following form:
\begin{equation} \label{message passing}
m_i^l = \bigoplus_{j \in \mathcal{N}(i)} m_{ij}(h^l(\textbf{x}_i), h^l(\textbf{x}_j), e_{ij}^l),    
\end{equation}
 and $\bigoplus_{j \in \mathcal{N}(i)}$ is any permutation-invariant pooling operation between the 1-hop neighbors of $i$. We also include the edge features $e_{ij}^l$ into the message passing phase for completeness. 3D \textbf{equivariant} MPNNs (3D GNNs for short) require the message $m_i$ to be equivariant with respect to the geometric graph. That is, for an arbitrary edge $e_{ij}$:
\begin{equation} \label{message}
m_{ij}(h^l(g\textbf{x}_i), h^l(g\textbf{x}_j)) = \mathcal{M}(g) m_{ij}(h^l(\textbf{x}_i), h^l(\textbf{x}_j)), \end{equation}
where $g \in SE(3)$ is acting on the whole geometric graph simultaneously: $(\textbf{x}_1, \dots, \textbf{x}_n) \rightarrow (g\textbf{x}_1, \dots, g\textbf{x}_n). $ For example, the invariant model ComENet~\cite{wangcomenet} satisfies Eq. \ref{message} by setting $\mathcal{M}(g) \equiv 1$, and MACE~\cite{batatiamace} realized Eq. \ref{message} for nonconstant irreducible group representations $\mathcal{M}(g)$ through spherical harmonics and Clebsch-Gordan coefficients.

\section{A Local Hierarchy of 3D Isomorphism}
\label{sec:local}

\begin{figure*}[t]
     \begin{center}
      %\vspace{-20 pt}
     \subfloat[] 
     {\includegraphics[width=0.48\textwidth]{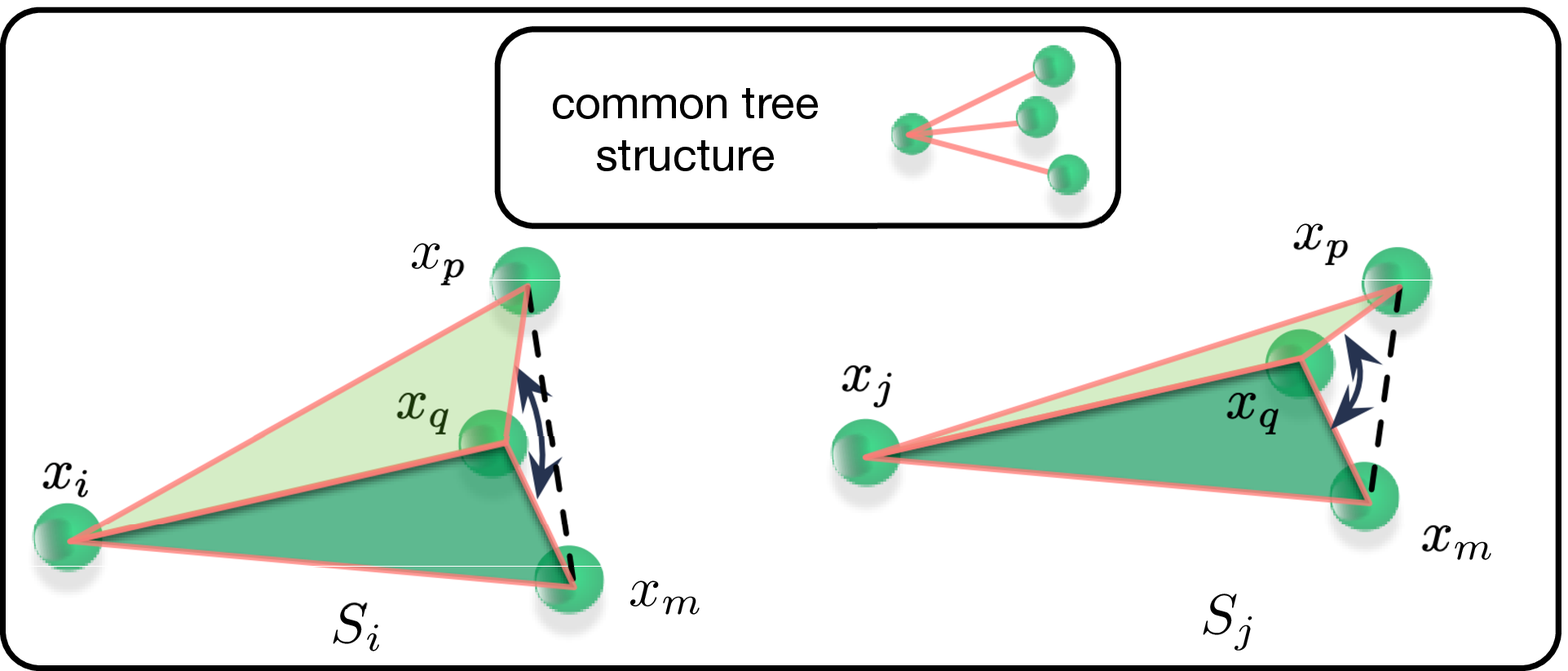}\label{fig:tree}}
     \quad
     \subfloat[] 
     {\includegraphics[width=0.48\textwidth]{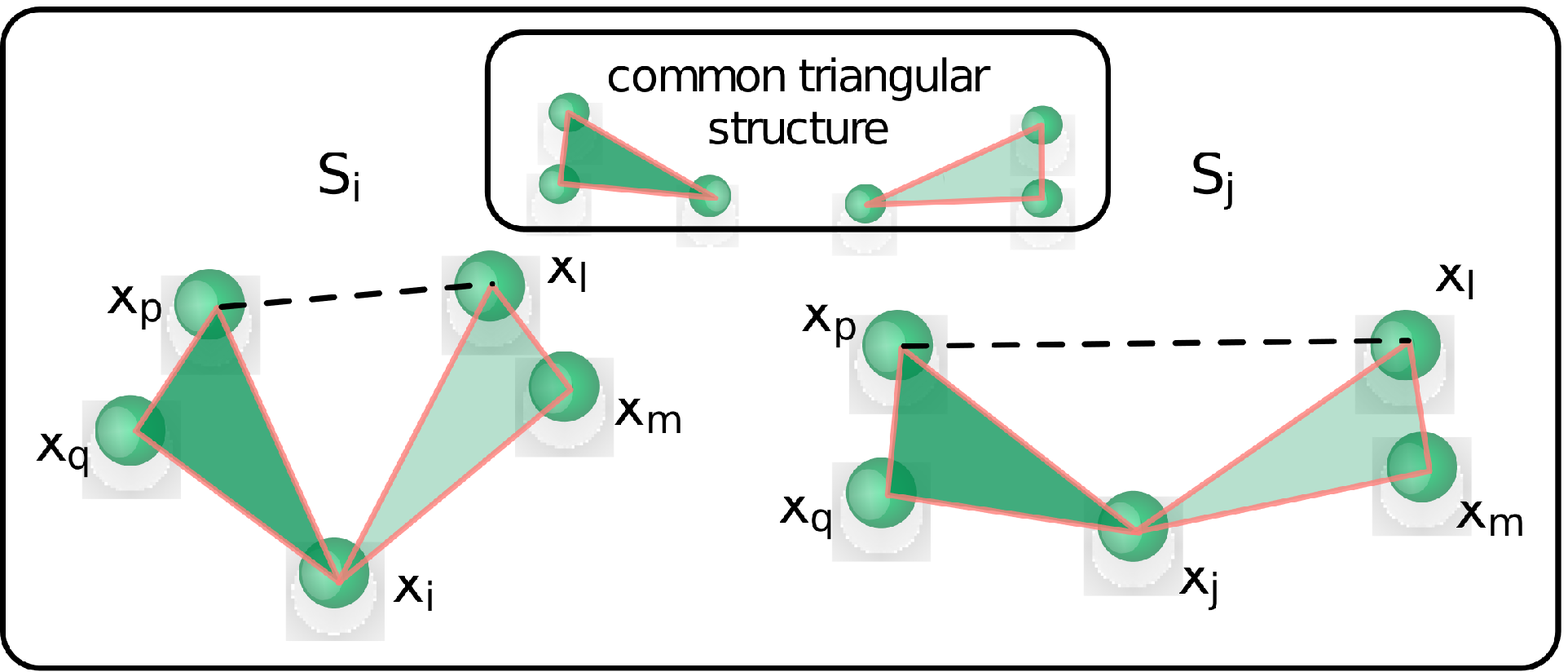}\label{fig:triangle}}
      \vspace{-10 pt}
    \caption{
    \protect\subref{fig:tree} $\textbf{S}_i$ and $\textbf{S}_j$ share the same tree structure (edge lengths are identical), but they are not triangular isomorphic (different dihedral angles); 
    \protect\subref{fig:triangle} $\textbf{S}_i$ and $\textbf{S}_j$ are triangular isomorphic but not subgraph isomorphic (the relative distance between the two triangles is different).
    }
    \vspace{-12pt}
    \label{fig:local_isomorphism}
    \end{center}
\end{figure*}

As presented in Section~\ref{sec:prelim}, defining expressive messages is an essential component for building powerful 3D GNNs. In this section, we develop a fine-grained characterization of local 3D structures and build its connection with the expressiveness of 3D GNNs.    

Since the celebrated work \cite{xu2018powerful}, a popular expressiveness test for permutation invariant graph neural networks is the 1-WL graph isomorphism test~\cite{morris2021weisfeiler}, and \citet{wijesinghe2021new} has shown that the 1-WL test is equivalent to the ability to discriminate the \textbf{local} subtree-isomorphism.
It motivates us to develop a novel (local) 3D isomorphism for testing the expressive power of 3D GNNs. However, this task is nontrivial, since most of the previous settings for graph isomorphism are only applicable to 2D topological features. For 3D geometric shapes, we should take the $SE(3)$ symmetry into account. Formally,
two 3D geometric graphs $\textbf{X},\textbf{Y}$ are defined to be \textbf{globally} isomorphic, if there exists $g \in SE(3)$ such that 
\begin{equation} 
\label{global}
\textbf{Y} = g \textbf{X}.
\end{equation}
In other words, $\textbf{X}$ and $\textbf{Y}$ are essentially the same, if they can be transformed into each other through a series of rotations and translations. Inspired by~\citet{wijesinghe2021new}, now we introduce a novel hierarchy of $SE(3)$ equivariant local isomorphism to measure the local similarity of 3D structures.

Let $\textbf{S}_i$ denote the 3D subgraph (and the associated node features) of node $i$, which contains all edges in $E$ if the end points are one-hop neighbors of $i$.  
For each edge $e_{ij} \in E$, the mutual 3D substructure $\textbf{S}_{i-j}$ is defined by the intersection of $\textbf{S}_i$ and $\textbf{S}_j$: $\textbf{S}_{i-j} = \textbf{S}_i \cap \textbf{S}_j$.

Given two local subgraphs $\textbf{S}_i$ and $\textbf{S}_j$ that correspond to two nodes $i$ and $j$, we say $\textbf{S}_i$ is $\{\text{-tree},\ \text{-triangular},\ \text{-subgraph}\}$ isometric to 
$\textbf{S}_j$, if there exists a bijective function $f: \textbf{S}_i \rightarrow \textbf{S}_j$ such that $h_{f(u)} = h_u$ for every node $u \in \textbf{S}_i$, and the following conditions hold respectively:
\begin{itemize}
    \item \textbf{Tree Isometric:} If there exists a collection of group elements $g_{iu} \in SE(3)$, such that $(\textbf{x}_{f(u)}, \textbf{x}_{f(i)}) = (g_{iu}\textbf{x}_u, g_{iu}\textbf{x}_i)$ for each edge $e_{iu} \in \textbf{S}_i$;
    \item \textbf{Triangular Isometric:} If there exists a collection of group elements $g_{iu} \in SE(3)$, such that the corresponding mutual 3D substructures satisfy: $\textbf{S}_{f(u)-f(i)} = g_{iu} \textbf{S}_{u-i}$ for each edge $e_{iu} \in \textbf{S}_{i-j}$;
    \item \textbf{Subgraph Isometric:} for any two adjacent nodes $u, v \in \textbf{S}_i$, $f(u)$ and $f(v)$ are also adjacent in $\textbf{S}_j$, and there exist a single group element $g_i \in SE(3)$ such that $g_i \textbf{S}_i = \textbf{S}_j$.

\end{itemize}
Note that tree isomorphism only considers edges around a central node, which is of a tree shape. On the other hand, the mutual 3D substructure can be decomposed into a bunch of triangles (since it's contained in adjacent node triplets), which explains the name of triangular isomorphism. 

In fact, the three isomorphisms form a hierarchy from micro to macro, in the sense that the following implication relation holds:
\begin{align*}
\textbf{Subgraph Isometric} \Rightarrow \textbf{Triangular Isometric}
\Rightarrow \textbf{Tree Isometric}
\end{align*}
This is an obvious fact from the above definitions. To deduce the reverse implication relation, we provide a visualized example. Figure \ref{fig:local_isomorphism} shows two examples of local 3D structures: 1. the first one shares the same tree structure, but is not triangular-isomorphic; 2. the second one is triangular-isomorphic but not subgraph-isomorphic. In conclusion, the following diagram holds:
\begin{align*}
\textbf{Tree Isometric} \not\Rightarrow \textbf{Triangular Isometric}
\not\Rightarrow  \textbf{Subgraph Isometric}
\end{align*}
One way to formally connect the expressiveness power of a geometric GNN with their ability of differentiating geometric subgraphs is to define geometric WL tests, the reader can consult \cite{joshi2022expressive}. In this paper, we take an intuitive approach based on our nested 3D hierarchy. That is, if two 3D GNN algorithms A and B can differentiate all non-isomorphic local 3D shapes of tree (triangular)  level, while A can differentiate at least two more 3D geometries which are non-isomorphic at triangular(subgraph) level than B, then we claim that algorithm A’s expressiveness power is more powerful than B.

Since tree isomorphism is determined by the one-hop Euclidean distance between neighbors, distinguishing local tree structures is relatively simple for ordinary 3D equivariant GNNs. For example, the standard baseline SchNet~\cite{schutt2018schnet} is one instance of Eq.~\ref{message passing} by setting $e^t_{ij} = \textbf{RBF}(d(\textbf{x}_i, \textbf{x}_j))$, 
where $\textbf{RBF}(\cdot)$ is a set of radial basis functions. Although it is powerful enough for testing tree non-isomorphism (assuming that $\textbf{RBF}(\cdot)$ is injective), we prove in Appendix \ref{appendix:isomor} that SchNet cannot distinguish non-isomorphic structures at the triangular level.

On the other hand, \citet{wijesinghe2021new} has shown that by leveraging the topological information extracted from local overlapping subgraphs, we can enhance the expressive power of GNNs to go beyond 2D sub-tree isomorphism. In our setting, the natural analogue of the overlapping subgraphs is exactly the mutual 3D substructures. Now 
we demonstrate how to merge the information from 3D substructures to the message passing framework (\ref{message passing}).
Given an $SE(3)$-\textbf{invariant} encoder $\phi$, define the 3D structure weights $A_{ij}: = \phi (\textbf{S}_{i-j})$ for each edge $e_{ij} \in E$. Then, the message passing framework (\ref{message passing}) is generalized to:
\begin{equation} \label{weighted message passing}
m_i^{l} = \bigoplus_{j \in \mathcal{N}(i)} m_{ij}(h^l(\textbf{x}_i), 
 h^l(\textbf{x}_j), A_{ij}h^l(\textbf{x}_j), e_{ij}^l).  
\end{equation}

Formula \ref{weighted message passing} is an efficient realization of enhancing 3D GNNs by injecting the mutual 3D substructures. However, a crucial question remains to be answered: \textit{Can the generalized message passing framework boost the expressive power of 3D GNNs?} Under certain conditions, the following theorem provides an affirmative answer:

\begin{theorem} \label{thm isomorphism}
Suppose $\phi$ is a a universal SE(3)-invariant approximator of functions with respect to the mutual 3d structures $\textbf{S}_{i-j}$, then the collection of weights $\{\{A_{ij}\}_{e_{ij} \in E}\}$ is able to differentiate local structures beyond tree isomorphism. Moreover, with additional injectivity assumptions (see Eq. \ref{injective}), 3D GNNs based on the enhanced message passing framework \ref{weighted message passing} map at least two distinct local 3D subgraphs with isometric local tree structures to different representations.
\end{theorem}
This theorem confirms that the enhanced 3D GNN (formula \ref{weighted message passing}) is more expressive than the SchNet baseline, at least in testing local non-isomorphic geometric graphs. The complete proof is left in Appendix \ref{appendix:isomor}. The existence of such local invariant encoder $\phi$ is also proved by explicit construction.
Note that there are other different perspectives on characterizing 3D structures, we will also briefly discuss them in Appendix \ref{appendix:isomor}.

\section{From Local to Global: The Missing Pieces}
\label{sec:build_express}

In the last section, we introduced a geometric local isomorphism hierarchy for testing the expressive power of 3D GNNs. Furthermore, we motivated adding a SE(3)-\textbf{invariant} encoder to improve the expressive power of one-hop 3D GNNs by scalarizing not only pairwise distances but also their mutual 3D structures in Theorem~\ref{thm isomorphism}. However, to build a powerful 3D GNN, it remains to be analyzed how a 3D GNN acquires higher order (beyond 1-hop neighbors) information by accumulating local messages. A natural question arises: 
\textit{are invariant features enough for representing \textbf{global} geometric information?}  

\begin{wrapfigure}{r}{0.5\textwidth}
    \includegraphics[width=0.48\textwidth]{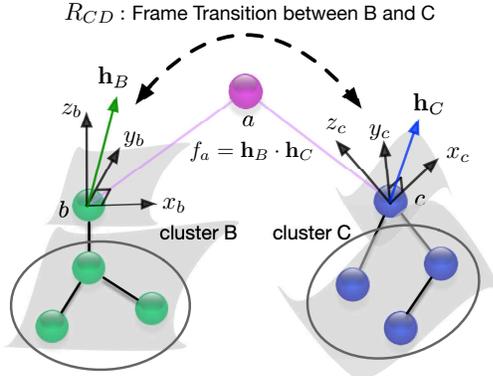}
    \caption{Illustrations of different local frames and their transition.}\label{fig:frame}
\end{wrapfigure}

To formally formulate this problem, we consider a two-hop aggregation case. From figure \ref{fig:frame}, the central atom $a$ is connected with atoms $b$ and $c$. Except for the common neighbor $a$, other atoms that connect to $b$ and $c$ form two 3D clusters, denoted by $\textbf{B}$, $\textbf{C}$. Suppose the ground-truth interaction potential of $\textbf{B}$ and $\textbf{C}$ imposed on atom $a$ is described by a tensor-valued function $f_{a}(\textbf{B},\textbf{C})$.
Since $\textbf{B}$ and $\textbf{C}$ are both beyond the 1-hop neighborhood of $a$, the information of $f_{a} (\textbf{B},\textbf{C})$ can only be acquired after two steps of message passing: 1. atoms $b$ and $c$ aggregate message separately from $\textbf{B}$ and $\textbf{C}$; 2. the central atom $a$ receives the aggregated message (which contains information of $\textbf{B}$ and $\textbf{C}$) from its neighbors $b$ and $c$. 

 Let $S_\textbf{B}$ ($S_\textbf{C}$) denote the collection of all invariant scalars created by $\textbf{B}$ ($\textbf{C}$) . For example, $S_\textbf{B}$ contains all relative distances and angles within the 3D structure $\textbf{B}$. Then, the following theorem holds: 
\begin{theorem} \label{localtoglobal}
Not all types of invariant interaction $f_{a}(\textbf{B},\textbf{C})$ can be expressed by inputting the union of two sets $S_\textbf{B}$ and $S_\textbf{C}$. In other words, there exists  $E(3)$ invariant function $f_{a}(\textbf{B},\textbf{C})$, such that it cannot be expressed as functions of $S_\textbf{B}$ and $S_\textbf{C}$:
$f_{a}(\textbf{B},\textbf{C}) \neq \rho(S_\textbf{B}, S_\textbf{C})$
for an arbitrary invariant function $\rho$. 
\end{theorem}
This theorem in essence tells us that naively aggregating `local' scalar information from different clusters is not enough to approximate `global' interactions, even if we only consider simple \textbf{invariant} interaction tasks. Different from the last section, where the local expressiveness is measured by the ability of classifying geometric shapes, we built regression functions that depend strictly more than the combination of local invariant scalars. 

Intuitively, the proof is based on the fact that all scalars in $S_\textbf{B}$ ($S_\textbf{C}$) can be expressed through  equivariant frames separately determined by $\textbf{B}$ ($\textbf{C}$). However, the transition matrix between these two frames is not encoded in the aggregation, which causes \textbf{information loss} when aggregating geometric features from two sub-clusters. More importantly, the proof also revealed the missing information that causes the expressiveness gap: Frame Transition (FT).

\textbf{Frame Transition (FT).} Formally, two orthonormal frames $(e^i_1, e^i_2, e^i_3)$ and $(e^j_1, e^j_2, e^j_3)$  are connected by an 
orthogonal matrix $R_{ij} \in SO(3)$:
\begin{equation} \label{transition}
(\textbf{e}^i_1, \textbf{e}^i_2, \textbf{e}^i_3) = R_{ij}(\textbf{e}^j_1, \textbf{e}^j_2, \textbf{e}^j_3).  \end{equation}
Moreover, it is easy to check that when $(\textbf{e}^i_1, \textbf{e}^i_2, \textbf{e}^i_3)$ and $(\textbf{e}^j_1, \textbf{e}^j_2, \textbf{e}^j_3)$ are equivariant frames, all elements of $R_{xy}$ are invariant scalars. Suppose $i$ and $j$ represent indexes of two connected atoms in a geometric graph, then the fundamental torsion angle $\tau_{ij}$ appeared in ComeNet~\cite{wangcomenet} is just one element of $R_{ij}$ (see Appendix~\ref{appendix:4}). 

Towards filling this expressiveness gap, we can straightforwardly inject all invariant pairwise frame transition matrices (\textbf{FT}) into the model. Nevertheless, it imposes expensive computational cost when the number of local clusters is large ($O(k^2)$ pairs of \textbf{FT} for each node). 
Therefore, compared with pure invariant approaches, a more efficient way is to introduce equivariant tensor features for each node $i$, denoted by $\textbf{m}_i$. By directly maintaining the equivariant frames in $\textbf{m}_i$, we show in Appendix~\ref{appendix:4} that  \textbf{FT} is easily derived through equivariant message passing.

\textbf{Equivariant Message Passing.} Similarly with the standard one-hop message passing scheme \ref{message passing}, the aggregated tensor message $\textbf{m}_i$ from the $l-1$ layer to the $l$ layer can be written as: $\textbf{m}_i^{l-1} = \sum_{j \in N(i)} \textbf{m}_j^{l-1}.$
Since summation does not break the symmetry rule, it is obvious that $\textbf{m}_i^{l-1}$ are still equivariant tensors. However, the nontrivial part lies in the design of the equivariant update function $\phi$:
\begin{equation} 
\label{vector update}
\textbf{m}_i^{l}= \phi (\textbf{m}_i^{l-1}).
\end{equation}
A good $\phi$ should have enough expressive power while preserving $SE(3)$ equivariance. Here, we propose a novel way of updating scalar and tensor messages by performing node-wise scalarization and tensorization blocks (the \textbf{FTE} module of Figure~\ref{fig:model}). From the perspective of Eq. \ref{message}, $\textbf{m}(\textbf{x}_u)$ is transformed equivariantly as:
\begin{equation} \label{equivariant agg}
 \textbf{m}(g\textbf{x}_u) = \sum_{i=0}^l \mathcal{M}^i(g)\textbf{m}_i(g\textbf{x}_u) ,\ \ g \in SE(3).
\end{equation}
Here, $\textbf{m}(\textbf{x}_u)$ is decomposed to $(\textbf{m}_0(\textbf{x}_u),  \dots, \textbf{m}_l(\textbf{x}_u))$ according to different tensor types, and $\{\mathcal{M}^i(g)\}_{i=0}^{l}$ is a collection of different $SE(3)$ \textbf{tensor representations} (see the precise definition in Appendix \ref{appendix:back}). 

To illustrate the benefit of aggregating equivariant messages from local patches, we study a simple case. Let $f_a(\textbf{B},\textbf{C}) = \textbf{h}_B \cdot \textbf{h}_C$ be an invariant function of $\textbf{B}$ and $\textbf{C}$ (see Fig. \ref{fig:frame}), then $f_a$ can be calculated by a direction composition of scalar messages and equivariant vector messages:
$f_a(\textbf{B},\textbf{C}) = \frac{1}{2}[\norm{\textbf{m}_a}^2 - \norm{\textbf{h}_B}^2 - \norm{\textbf{h}_C}^2],$
where $\textbf{m}_a = \textbf{h}_B + \textbf{h}_C$ is an equivariant vector. Note that $\textbf{m}_a$ follows the local equivariant aggregation formula \ref{vector update}, and the other vectors' norm $\norm{\textbf{h}_B}$ and $\norm{\textbf{h}_C}$  are obtained through local scalarization on atoms $b$ and $c$. As a comparison, it's worth mentioning that $f_a(\textbf{B},\textbf{C})$ can also be expressed by local scalarization with the additional transition matrix data $R_{BC}$ defined by Eq. \ref{transition}. Let $\Tilde{h}_B$ and $\Tilde{h}_C$ be the scalarized coordinates with respect to two local equivariant frames $\mathcal{F}_{B}$ and $\mathcal{F}_{C}$. Then 
$f_a(\textbf{B},\textbf{C}) = \frac{1}{2}\left[\norm{R_{BC}^{-1}\Tilde{h}_B + \Tilde{h}_C}^2 - \norm{\Tilde{h}_B}^2 - \norm{\Tilde{h}_C}^2\right].$
However, it requires adding the rotation matrix $R_{BC}$ for each $(\textbf{B},\textbf{C})$ pair, which is computationally expensive compared to directly implementing equivariant tensor updates.

\section{Building an Efficient and Expressive Equivariant 
 3D GNN}
\label{sec:leftnet}

\begin{figure*}[t]
    %\vspace{-12 pt}
    \centering
    \includegraphics[width=0.98\textwidth]{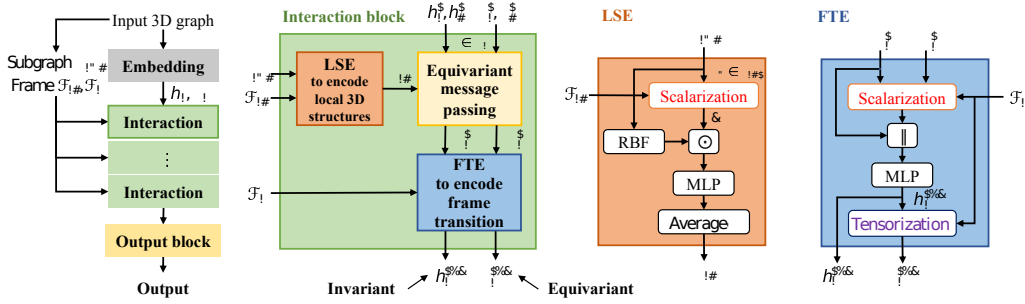}
    \vspace{-5 pt}
    \caption{Illustrations of our modular framework for building equivariant GNNs and the realization of LEFTNet. Each interaction block contains \textbf{LSE} to encode local 3D structures, equivariant message passing to update both invariant (unbold letters, e.g. $h_i$) and equivariant (\textbf{bold} letter, e.g. $\mathbf{h}_i$) features, and \textbf{FTE} to encode frame transition. $\textbf{S}_{i-j}$ is the local 3D structure of each edge $e_{ij}$. $\mathcal{F}_{ij}$ and $\mathcal{F}_i$ are the equivariant frames for each edge $e_{ij}$ and node $i$. $\odot$ indicates element-wise multiplication, and $\|$ indicates concatenation. Note that we do not include $\mathbf{e}_{ij}$ in the figure since, practically, they are generated based on $\mathbf{h}_i$ and $\mathbf{h}_j$.}
    \label{fig:model}
    \vspace{-10 pt}
\end{figure*}

We propose to leverage the full power of \textbf{LSE} and \textbf{FTE} along with a powerful \textbf{tensor update} module to push the limit of efficient and expressive 3D equivariant GNNs design.

\textbf{LSE Instantiation.}
We propose to apply edge-wise equivariant frames to encode the local 3D structures $\textbf{S}_{i-j}$. By definition, $\textbf{S}_{i-j}$ contains edge $e_{ij}$, nodes $i$ and $j$, and their common neighbors. We use the equivariant frame $\mathcal{F}_{ij}$ built on $e_{ij}$ (see the precise formula in Appendix \ref{appendix:5}) to scalarize $\textbf{S}_{i-j}$.
After scalarization (\ref{scalarization}), the equivariant coordinates of all nodes in $\textbf{S}_{i-j}$ are transformed into invariant coordinates: $\{\textbf{x}_k \rightarrow \Tilde{x}_k\ \text{for}\ \textbf{x}_k \in \textbf{S}_{i-j}\}$. To encode these scalars sufficiently, we first weight each $\Tilde{x}_k$ by the \textbf{RBF} distance embedding: $\Tilde{x}_k \rightarrow \textbf{RBF}(\norm{\textbf{x}_k}) \odot \text{MLP}(\Tilde{x}_k)$ for each $\textbf{x}_k \in \textbf{S}_{i-j}$.
Note that to preserve the permutation symmetry, the \text{MLP} is shared among the nodes. Finally, the 3D structure weight $A_{ij}$ is obtained by the average pooling of all node features.  

\textbf{FTE Instantiation.}
We propose to introduce equivariant tensor message passing and update function for encoding local \textbf{FT} information. At initialization, let $\textbf{NF}^l(\textbf{x}_i,\textbf{x}_j)$ denote the embedded tensor-valued edge feature between $i$ and $j$. We split it into two parts: 1. the scalar part $\text{SF}^l(\textbf{x}_i,\textbf{x}_j)$ for aggregating invariant messages; 2. the higher order tensor part $\textbf{TF}^l(\textbf{x}_i,\textbf{x}_j)$ for aggregating tensor messages. To transform $\textbf{TF}^l(\textbf{x}_i,\textbf{x}_j)$, we turn to the equivariant frame $\mathcal{F}_{ij}$ once again. After scalarization by $\mathcal{F}_{ij}$, $\textbf{TF}^l(\textbf{x}_i,\textbf{x}_j)$ becomes a tuple of scalars $\Tilde{\text{TF}}^l(\textbf{x}_i,\textbf{x}_j)$, which is then transformed by MLP. Finally, we output arbitrary tensor messages through equivariant tensorization \ref{higher tensorization}: $$\Tilde{\text{TF}}^l(\textbf{x}_i,\textbf{x}_j) \xrightarrow[\mathcal{F}_{ij}]{\textbf{Tensorize}} \textbf{NF}^{l+1}(\textbf{x}_i,\textbf{x}_j).$$
Further details are provided in Appendix \ref{appendix:5}. As we have discussed earlier, the node-wise tensor update function $\phi$ in Eq.~\ref{vector update} is also one of the guarantees for a powerful \textbf{FTE}. As a comparison, $\phi$ is usually a standard MLP for updating node features in 2D GNNs, which is a \textbf{universal approximator} of invariant functions. Previous works~\cite{satorras2021n,jing2020learning} updated equivariant features by taking linear combinations and calculating the invariant norm of tensors, which may suffer from information loss.  
Then a natural question arises: \textit{Can we design an equivariant universal approximator for tensor update?} We answer this question by introducing a novel node-wise frame. Consider node $i$ with its position $\textbf{x}_i$, let $\Bar{\textbf{x}}_i : = \frac{1}{N}\sum_{\textbf{x}_j \in N(\textbf{x}_i)} \textbf{x}_j$ 
be the center of mass around $\textbf{x}_i$'s neighborhood. Then the orthonormal equivariant frame $\mathcal{F}_i : = (\textbf{e}^i_1,\textbf{e}^i_2,\textbf{e}^i_3)$ with respect to $\textbf{x}_i$ is defined by
\begin{equation} 
\label{eq:node frame}
(\frac{\textbf{x}_i - \Bar{\textbf{x}}_i}{\norm{\textbf{x}_i - \Bar{\textbf{x}}_i}}, \frac{\Bar{\textbf{x}}_i \times \textbf{x}_i}{\norm{\Bar{\textbf{x}}_i \times \textbf{x}_i}},\frac{\textbf{x}_i - \Bar{\textbf{x}}_i}{\norm{\textbf{x}_i - \Bar{\textbf{x}}_i}} \times \frac{\Bar{\textbf{x}}_i \times \textbf{x}_i}{\norm{\Bar{\textbf{x}}_i \times \textbf{x}_i}}  ).
\end{equation}
Finally, we realize a powerful $\phi$ by the following theorem:
\begin{theorem} 
\label{thm vector update}
Equipped with an equivariant frame $\mathcal{F}_i$ for each node $i$, the equivariant function $\mathbf{\phi}$ defined by the following composition is a universal approximator of tensor transformations: $\mathbf{\phi}: \textbf{Scalarization}
 \rightarrow \textbf{MLP} \rightarrow \textbf{Tensorization}.$
\end{theorem}
The proof is left in Appendix \ref{appendix:5}.

\textbf{LEFTNet.}
An overview of our $\{\textbf{LSE}, \textbf{FTE}\}$ enhanced efficient graph neural network (LEFTNet) is depicted in Figure~\ref{fig:model}. LEFTNet receives as input a collection of node embeddings $\{v^0_1,\dots,v^0_N\}$, which contain the atom types and 3D positions for each node: $v^0_i = (z_i, \textbf{x}_i)$, where $i \in \{1,\dots,N\}$. For each edge $e_{ij} \in E$, we denote the associated equivariant features consisting of tensors by $\mathbf{e}_{ij}$. During each messaging passing layer, the \textbf{LSE} module outputs the scalar weight coefficients $A_{ij}$ as enhanced invariant edge feature and feed into the interaction module. Moreover, scalarization and tensorization as two essential blocks are used in the equivariant update module that fulfills the function of \textbf{FTE}. 
The permutation equivariance of a geometric graph is automatically guaranteed for any message passing architecture, we provide a complete proof of SE(3)-equivariance for LEFTNet in Appendix \ref{appendix:5}.

\textbf{SE(3) vs E(3) Equivariance.}
Besides explicitly fitting the $SE(3)$ invariant molecular geometry probability distribution, modeling the energy surface of a molecule system is also a crucial task for molecule property prediction.
However, the Hamiltonian energy function $E$ of a molecule is invariant under refection transformation:
$\textbf{Energy}(\mathbf{X}) = \textbf{Energy}(R\mathbf{X}),$
for arbitrary reflection transformation $R \in E(3)$.
In summary, there exist two different inductive biases for modeling 3D data: \textbf{(1)} SE(3) equivariance, e.g. chirality could turn a therapeutic drug to a killing toxin; \textbf{(2)} E(3) equivariance, e.g. energy remains the same under reflections.

Since we implement $SE(3)$ equivariant frames in LEFTNet, our algorithm is naturally $SE(3)$ equivariant (and reflection anti-equivariant). However, our method is \textbf{flexible} to implement $E(3)$ equivariant tasks as well.
For $E(3)$ equivariance, we can either replace our frames to $E(3)$ equivariant frames, or  modify the scalarization block by taking the absolute value:
$\textbf{x} \rightarrow \Tilde{x}: = \underbrace{(\textbf{x}\cdot e_1,\textbf{x}\cdot e_2, \textbf{x}\cdot e_3)}_{SE(3)} \rightarrow \underbrace{(\textbf{x}\cdot e_1,|\textbf{x}\cdot e_2|, \textbf{x}\cdot e_3)}_{E(3)}.$
Intuitively, since the second vector $e_2$ is a pseudo-vector, projections of any equivariant vectors along the $e_2$ direction are not $E(3)$ invariant until taking the absolute value.

 \textbf{Efficiency.}
To analyze the efficiency of LEFTNet, suppose 3D graph $G$ has $n$ vertices, and its average node degree is $k$. Our algorithm consists of three phases: 1. Building equivariant frames and performing local scalarization;  2. Equivariant message passing; 3. Updating node-wise tensor features through scalarization and tensorization. Let $l$ be the number of layers, then the computational complexity for each of our three phases are: 1. $O(nk)$ for computing the frame and local (1-hop) 3D features; 2. $O(nkl)$ for 1-hop neighborhood message aggregation; 3. $O(nl)$ for node-wise tensorization and feature update.

\section{Related Work}

\begin{table}[t]  
\caption{Categorization of representative geometric GNN algorithms. $^*$ denotes partially satisfying the requirement.}\label{tab:cat_related_works}
% \scriptsize
\centering
\resizebox{0.7\linewidth}{!}{
\begin{tabular}{lcccccc}\toprule
\textbf{Method} &  \textbf{Symmetry} & \textbf{LSE} &\textbf{FTE} &\textbf{Complexity} \\
\midrule
SchNet~\cite{schutt2018schnet}       & E(3)-invariant         & \xmark     & \xmark        & $O(nk)$ \\
EGNN~\cite{satorras2021n}            & E(3)-equivariant       & \xmark     & \cmark$^*$    & $O(nk)$ \\
GVP-GNN~\cite{jing2020learning}      & E(3)-equivariant       & \xmark     & \cmark        & $O(nk)$ \\
ClofNet~\cite{du2022se}              & SE(3)-equivariant      & \xmark     & \xmark        & $O(nk)$ \\
PaiNN~\cite{schutt2021equivariant}   & E(3)-equivariant       & \xmark     & \cmark        & $O(nk)$ \\
ComENet~\cite{wangcomenet}           & SE(3)-invariant        & \cmark     & \cmark$^*$    & $O(nk)$ \\
TFN~\cite{thomas2018tensor}          & SE(3)/E(3)-equivariant & \xmark     & \cmark        & $O(nk)$ \\
Equiformer~\cite{liao2022equiformer} & SE(3)/E(3)-equivariant & \xmark     & \cmark        & $O(nk)$ \\
% NequIP~\cite{batzner20223} & SE(3)/E(3)-equivariant & \xmark & \cmark & $O(nk)$ \\
SphereNet~\cite{liu2021spherical}    & SE(3)-invariant        & \cmark$^*$ & \cmark$^*$    & $O(nk^2)$ \\
GemNet~\cite{gasteiger2021gemnet}    & SE(3)-invariant        & \cmark$^*$   & \cmark$^*$    & $O(nk^3)$ \\
\midrule
LEFTNet (Ours)                       & SE(3)/E(3)-equivariant & \cmark     & \cmark        & $O(nk)$ \\
\bottomrule
\end{tabular}
}
\end{table}

In light of the discussions in Section~\ref{sec:local} and~\ref{sec:build_express}, we summarize two necessary ingredients for building expressive equivariant 3D GNNs: (1) local 3D substructure encodings (\textbf{LSE}), such that the local message is aware of different local 3D structures; (2) frame transition encodings (\textbf{FTE}), such that the 3D GNN is aware of the equivariant coordinate transformation between different local patches.
 
We review the previous 3D GNNs following this framework and summarize the results in Table~\ref{tab:cat_related_works}. For a fair comparison, we also list the computational complexity as it is often a trade-off of expressiveness (see the detailed analysis at the end of the next section). For \textbf{LSE}, SphereNet~\cite{liu2021spherical} and GemNet~\cite{gasteiger2021gemnet} (implicitly) encode the local 3D substructures by introducing a computation-intensive 2-hop edge-based update. For \textbf{FTE}, most 3D GNNs with equivariant vector update are able to express the local frame transitions (\textbf{FT}). While EGNN~\cite{satorras2021n} is an exception, because it only updates the position vector (i.e. one channel), which is insufficient to express the whole \textbf{FT}. In other words, whether the update function $\phi$ of (\ref{vector update}) is powerful also affects the \textbf{FT} encoding. Except for equivariant update methods, models that encode torsion angle information also partially express \textbf{FTE} as illustrated in Appendix \ref{appendix:4}. However, there is a trade-off between the efficiency and expressiveness in terms of number of hops considered for message passing.   

Different from our invariant realization of \textbf{LSE}, \citet{batatiamace} builds its framework by constructing complete equivariant polynomial basis with the help of spherical harmonics and tensor product, where the monomial variables depend on different nodes (bodies). On the other hand,  we realize the function of \textbf{LSE} and \textbf{FTE} through the edgewise scalarization $A_{ij}$ and the equivariant message passing (see Fig.~\ref{fig:model}). 

Recently, ~\citet{joshi2022expressive} propose a \textbf{geometric k-WL test} (GWL) to measure the expressiveness power of geometric GNN algorithms. On a high level, our tree isomorphism is equivalent to the 1-hop geometric isomorphism as proposed in GWL, and the fine-grained triangular isomorphism lies between the 1-hop and 2-hop geometric isomorphism as proposed in GWL. From the model design point of view, our realization of \textbf{LSE} is through local scalarization, whose expressiveness is guaranteed by the Kolmogorov representation theorem (see~\cite{zaheer2017deep}) and the universal approximator property of MLP. Moreover, the key concepts of measuring the expreesive power in ~\cite{joshi2022expressive} are the body order and tensor order, which originate from classical inter-atomic potential theories and are of the equivariance nature. On the other hand, we discover the \textbf{FTE} as the 'missing' bridge connecting local invariant scalars and global geometric expressiveness, which (together with \textbf{LSE} on mutual 3D substructures) also reveals why the 1-hop scalarization implemented in ClofNet~\cite{du2022se} is insufficient.

\section{Experiments}

\begin{table*}[t]
  \centering
  \caption{Mean Absolute Error for the molecular property prediction benchmark on QM9 dataset. (The best results are \textbf{bolded} and the second best are \underline{underlined}.)
  }
  \resizebox{0.95\textwidth}{!}{
  \begin{tabular}{l c c c c c c c c c c c c}
  \toprule
    Task & $\alpha$ & $\Delta \varepsilon$ & $\varepsilon_{\mathrm{HOMO}}$ & $\varepsilon_{\mathrm{LUMO}}$ & $\mu$ & $C_{\nu}$ & $G$ & $H$ & $R^2$ & $U$ & $U_0$ & ZPVE \\
    Units & bohr$^3$ & meV & meV & meV & D & cal/mol K & meV & meV & bohr$^3$ & meV & meV & meV \\
    \midrule
    NMP & .092 & 69 & 43 & 38 & .030 & .040 & 19 & 17 & .180 & 20 & 20 & 1.50 \\
    Cormorant & .085 & 61 & 34 & 38 & .038 & .026 & 20 & 21 & .961 & 21 & 22 & 2.03  \\
    LieConv & .084 & 49 & 30 & 25 & .032 & .038 & 22 & 24 & .800 & 19 & 19 & 2.28 \\
    TFN & .223 & 58 & 40 & 38 & .064 & .101 & - & - & - & - & - & - \\
    SE(3)-Tr. & .142 & 53 & 35 & 33 & .051 & .054 & - & - & - & - & - & -  \\
    EGNN & .071  & 48 & 29 & 25 & .029 & .031 & 12 & 12 & \textbf{.106} & 12 & 11 & 1.55  \\
    SEGNN &  .060& 42& 24& 21& .023& .031& 15& 16& .660& 13& 15& 1.62\\
    ClofNet & .063 & 53 & 33 & 25 & .040 & .027 & \underline{9} & \underline{9} & .610 & \underline{9} & \underline{8} & \textbf{1.23} \\
    EQGAT & .063& 44 &26& 22& \underline{.014} & .027& 12& 13 &.257& 13 &13 &1.50\\
    Equiformer & \underline{.056} & \textbf{33} & \textbf{17} & \textbf{16} & \underline{.014} & \underline{.025} & 10 &10 &.227 &11 &10 &\underline{1.32}\\
    LEFTNet (ours) & \textbf{.048} & \underline{40} & \underline{24} & \underline{18} & \textbf{.012} & \textbf{.023}
 & \textbf{7} & \textbf{6} & \underline{.109} & \textbf{7} & \textbf{6} & 1.33\\
    \midrule
    Schnet & .235 & 63 & 41 & 34 & .033 & .033 & 14 & 14 & \underline{.073} & 19 & 14 & 1.70 \\
    DimeNet++ & .044 & \underline{33} & 25 & 20 & .030 & .023 & 8 & 7 & .331 & 6 & \underline{6} & 1.21  \\
    SphereNet & .046 & \textbf{32} & \textbf{23} & \textbf{18} & .026 & \textbf{.021} & 8 & \underline{6} & .292 & 7 & \underline{6} & \textbf{1.12}\\
     ClofNet & .053 & 49 & 33 & 25 & .038 & .026 & 9 & 8 & .425 & 8 & 8 & 1.59\\
    PaiNN & .045 & 46 & 28 & 20 & \underline{.012} & .024 & \underline{7} & \underline{6} & \textbf{.066} & 6 & \underline{6} & 1.28\\
    LEFTNet (ours) & \underline{.039} & 39 & \textbf{23} & \textbf{18} & \textbf{.011} & \underline{.022} & \textbf{6} & \textbf{5} & .094 & \textbf{5} & \textbf{5} & \underline{1.19} \\
    \bottomrule
  \end{tabular}}
  \vspace{-10 pt}
  \label{tab:exp_qm9}
\end{table*}

\begin{table*}[b]
\centering 
\vspace{-10 pt}
\caption{Mean Absolute Error for per-atom forces prediction (kcal/mol \AA) on MD17 dataset. Baseline results are taken from the original papers (with unit conversions if needed). All models are trained on energies and forces, and WoFE is the weight of force over energy in loss functions. The best results are \textbf{bolded}.
}\label{tab:exp_md17}
\resizebox{0.99\textwidth}{!}{
\begin{tabular}{lcccccc|ccc|cc}\toprule
&\multicolumn{6}{c|}{WoFE=100} &\multicolumn{3}{c|}{WoFE=1000} &\multicolumn{2}{c}{Others}\\\midrule
Molecule &sGDML &SchNet &DimeNet &SphereNet &SpookyNet &LEFTNet &SphereNet &GemNet &LEFTNet &PaiNN &NewtonNet \\\midrule
Aspirin         &0.68 &1.35 &0.499 &0.430 &0.258 &\textbf{0.210} &0.209 &0.217 &\textbf{0.196} &0.371 &0.348 \\
Benzene         &0.20 &0.31 &0.187 &0.178 &--    &\textbf{0.145} &0.147 &0.145 &\textbf{0.142} &--    &--   \\
Ethanol         &0.33 &0.39 &0.230 &0.208 &\textbf{0.094} &0.118 &0.091 &\textbf{0.086} &0.099 &0.230 &0.264 \\
Malonaldehyde   &0.41 &0.66 &0.383 &0.340 &0.167 &\textbf{0.159} &0.172 &0.155 &\textbf{0.142} &0.319 &0.323 \\
Naphthalene     &0.11 &0.58 &0.215 &0.178 &0.089 &\textbf{0.063} &0.048 &0.051 &\textbf{0.044} &0.083 &0.084 \\
Salicylic acid  &0.28 &0.85 &0.374 &0.360 &0.180 &\textbf{0.141} &\textbf{0.113} &0.125 &0.117 &0.209 &0.197 \\
Toluene         &0.14 &0.57 &0.210 &0.155 &0.087 &\textbf{0.070} &0.054 &0.060 &\textbf{0.049} &0.102 &0.088 \\
Uracil          &0.24 &0.56 &0.301 &0.267 &0.119 &\textbf{0.117} &0.106 &0.097 &\textbf{0.085} &0.140 &0.149 \\
\bottomrule
\end{tabular}}
\end{table*}

We test the performance of LEFTNet on both scalar value (e.g. energy) and vector value (e.g. forces) prediction tasks. The scalar value prediction experiment is conducted on the QM9 dataset~\cite{ramakrishnan2014quantum} which includes $134k$ small molecules with quantum property annotations; the vector value prediction experiment is conducted on the MD17 dataset~\cite{chmiela2017machine} and the Revised MD17(rMD17) dataset~\cite{christensen2020role} which includes the energies and forces of molecules. We compare our LEFTNet with a list of state-of-the-art equivariant (invariant) graph neural networks including SphereNet~\cite{liu2021spherical}, PaiNN~\cite{schutt2021equivariant}, Equiformer~\cite{liao2022equiformer}, GemNet~\cite{gasteiger2021gemnet}, etc~\cite{schutt2018schnet,gasteiger2020directional,thomas2018tensor,fuchs2020se,anderson2019cormorant,satorras2021n,gilmer2017neural,finzi2020generalizing,le2022equivariant, musaelian2023learning, kovacs2021linear, faber2018alchemical, bartok2010gaussian, gao2020torchani}.The results on rMD17 and ablation studies are listed in Appendix~\ref{appendix:exp}.

\subsection{QM9 - Scalar-valued Property Prediction}
The QM9 dataset is a widely used dataset for predicting molecular properties. 
However, existing models are trained on different data splits. Specifically, Cormorant~\cite{anderson2019cormorant}, EGNN~\cite{satorras2021n}, etc., use 100k, 18k, and 13k molecules for training, validation, and testing, while DimeNet~\cite{gasteiger2020directional}, SphereNet~\cite{liu2021spherical}, etc., split the data into 110k, 10k, and 11k. For a fair comparison with all baseline methods, we conduct experiments using both data splits. Experimental results are listed in Table~\ref{tab:exp_qm9}.
For the first data split, LEFTNet is the best on 7 out of the 12 properties and improves previous SOTA results by 20\% on average. In addition, LEFTNet is the second best on 4 out of the other 5 tasks. 
Consistently, LEFTNet is the best or second best on 10 out of the 12 properties for the second split. These experimental results on both splits validate the effectiveness of LEFTNet on scalar-valued property prediction tasks.
The ablation study in Appendix~\ref{appendix:exp} shows that both \textbf{LSE} and \textbf{FTE} contribute to the final performance.

\subsection{MD17 - Vector-valued Property Prediction}
We evaluate the ability of LEFTNet to predict forces on the MD17 dataset. Following existing studies~\cite{schutt2018schnet, gasteiger2020directional, liu2021spherical}, we train a separate model for each of the 8 molecules. Both training and validation sets contain 1000 samples, and the rest are used for testing. Note that all baseline methods are trained on a joint loss of energies and forces, but different methods use different weights of force over energy (WoFE). For example, SchNet~\cite{schutt2018schnet} sets WoEF as 100, while GemNet~\cite{gasteiger2021gemnet} uses a weight of 1000. For a fair comparison with existing studies, we conduct experiments on two widely used weights of 100 and 1000 following~\citet{liu2021spherical}. The results are summarized in Table~\ref{tab:exp_md17}.  
We can observe that when WoFE is 100, LEFTNet outperforms all baseline methods on 7 of the 8 molecules and improves previous SOTA results by 16\% on average. In addition, LEFTNet can outperform all baseline methods on 6 of the 8 molecules when WoFE is 1000. These experimental results on MD17 demonstrate the performance of LEFTNet on vector-valued property prediction tasks.
The ablation study in Appendix~\ref{appendix:exp} also demonstrates that both \textbf{LSE} and \textbf{FTE} are important to the final results.

\section{Limitation and Future Work}
In this paper, we seek a general recipe for building 3D geometric graph deep learning algorithms. Considering common prior of 2D graphs, such as permutation symmetry, has been incorporated in off-the-shelf graph neural networks, we mainly focus on the $E(3)$ and $SE(3)$ symmetry specific to 3D geometric graphs. Despite our framework being general for modeling geometric objects, we only conducted experiments on commonly used molecular datasets. It’s worth exploring datasets in other domains in the future.

To elucidate the future design space of equivariant GNNs, we propose two directions that are worth exploring. Firstly, our current algorithms consider fixed equivariant frames for performing aggregation and node updates. Inspired by the high body-order ACE approach~\cite{batatia2022design} (for modeling atom-centered potentials), it is worth investigating in the future if  equivariant frames that relate to many body (e.g., the PCA frame in~\cite{puny2021frame}) can boost the performance of our algorithm. For example, to build the A-basis proposed in~\citet{puny2021frame}, we can replace our message aggregation Eq. \ref{vector update} from summation to tensor product, which is also a valid pooling operation. Another direction is to explore geometric mesh graphs on manifolds $M$, where the local frame is defined on the tangent space of each point: $\mathcal{F}(x) \in T_x M$. Since our scalarization technique (crucial for realizing \textbf{LSE} in LEFTNet) originates from differential geometry on frame bundles~\cite{hsu2002stochastic}, it is reasonable to expect that our framework also works for manifold data~\cite{he2021gauge,huang2022equivariant}.

\clearpage
\bibliographystyle{unsrtnat}
\bibliography{reference}

\newpage
\appendix
\newpage
\appendix
\onecolumn

\begin{center}
    \Large{
    Appendix}
\end{center}

\section{Supplementary Background}
\label{appendix:back}
We briefly review the concept of (contravariant) tensor fields and their associated equivariant group representations. 

A $s$ order (contravariant) tensor $\textbf{T}$ on a vector space $\mathbf{V}$ is a multilinear map:
$$T: \underbrace{\mathbf{V}^* \times \cdots \times \mathbf{V}^*}_s \rightarrow \mathbf{R}^1,$$
where $\mathbf{V}^*$ denotes the dual space of $\mathbf{V}$. In fact, there is a canonical `multiplication' operation between two tensors. Define the \textbf{tensor} \textbf{product} $\textbf{S} \otimes \textbf{T}$ of two tensors $\textbf{S}$ and $\textbf{T}$ to be a tensor of order $r+s$ :
\begin{equation} \label{def:tensor}
\textbf{S}\otimes \textbf{T} (v^1,\dots,v^{r+s}) = \textbf{S}(v^1,\dots,v^r) \textbf{T} (v^{r+1},\dots,v^{r+s}).\end{equation}
where $v^i \in \mathbf{V}^*$.

From now on, we assume $V = V^* = \mathbf{R}^3$. Note that when $s=1$, $\textbf{T}$ is exactly an equivariant vector. In practice, the tensor data in $\mathbf{R}^3$ is usually given by its coefficients under a Cartesian coordinate system. Take a second-order tensor as an example, assume we are given an orthonormal frame (basis) $(\textbf{e}_1, \textbf{e}_2, \textbf{e}_3)$ and its dual frame $(\textbf{e}^1, \textbf{e}^2, \textbf{e}^3)$, then the nine coefficients of $T$ are given by
$$T_{ij} = \textbf{T}(\textbf{e}^i,\textbf{e}^j),\ \ 1 \leq i,j \leq 3.$$
In other words, we say the collection $\{T_{ij}\}_{1 \leq i,j \leq 3}$ is a faithful representation of $\textbf{T}$ in a fixed coordinate system:
\begin{equation}     \label{second tensor}
\textbf{T} = \sum_{i,j}T_{ij} \textbf{e}_i \otimes \textbf{e}_j. \end{equation}
Once defined a tensor on $\mathbf{R}^3$, it's easy to extend it to a continuous manifold or a discrete graph. A \textbf{tensor field} of order $s$ on a 3D graph $G = (V,E)$ is a tensor-valued function $f$ which assigns to each 3D node $\textbf{x}_i$ an order $s$ tensor, denoted by $f(\textbf{x}_i)$.

\textbf{SE(3) Tensor Representations.} Let $V$ be a vector space, then the group SE(3) is said to act on $V$ if there is a mapping $\phi: SE(3) \times V \rightarrow V$ satisfying the following two conditions:
\begin{enumerate}
    \item if $e \in SE(3)$ is the identity element, then
    $$\phi(e,x) = x\ \ \ \ \text{for}\ \ \forall x \in V.$$
    \item if $g_1,g_2 \in SE(3)$, then
    $$\phi(g_1,\phi(g_2,x)) = \phi(g_1g_2, x)\ \ \ \ \text{for}\ \ \forall x \in V.$$
\end{enumerate}
If we further require $\phi(g,\cdot)$ is a linear map for all $g \in SE(3)$, then $\phi$ becomes a \text{group representation} of $SE(3)$. From now on, we only consider the rotation subgroup SO(3) and its group representations.  When $V = \mathbf{R}^3$, there is a natural representation of $SO(3)$ by rotating vectors in $\mathbf{R}^3$. In this way, an element $g \in SO(3)$ is identified with a Rotation matrix, denoted by $\{g_i^j\}_{1 \leq i,j \leq 3}.$

From the tensor definition (\ref{def:tensor}), this natural representation on $\mathbf{R}^3$ induces a tensor representation on $T$. Still take $\textbf{T} = \{T_{ij}\}_{1 \leq i,j \leq 3}$ as an example, we have
\begin{equation} \label{def:tensor rep}
T_{kl} = \sum_i \sum_j g_k^i g_l^j T_{ij},\ \ \ \ 1 \leq k,l \leq 3,    
\end{equation}
for $\forall g \in SO(3)$.
It's easy to check that (\ref{def:tensor rep}) is indeed a SO(3) representation on the vector space spanned by second-order tensors. 

\textbf{Relation with Spherical Harmonics.} For the $SO(3)$ group, all representations (including the tensor representations) can be decomposed as a direct sum of irreducible representations. For each type of irreducible representations, there is a subset of spherical harmonics formulating a basis for this specific representation. However, in terms of representing equivariant geometric quantities, the theorem in \cite{dym2020universality} claims that tensor representations and irreducible representations are equally powerful: They all form a complete basis in the space of continuous $E(3)$ equivariant functions.

\begin{algorithm}[tb]
   \caption{Invariant Design for LEFTNET.}
   \label{alg:invariant}
\begin{algorithmic}[1]
   \STATE {\bfseries Input:} Complete 3D gragh with equivariant positions $\mathbf{X}=(\textbf{x}_1,\dots,\textbf{x}_n) \in \mathbb{R}^{n\times 3}$, invariant node features $h_i \in \mathbb{R}^{d}$, invariant relative distances $d_{ij} \in \mathbb{R}^{1}$. 
   \STATE Centralize the positions: $\textbf{X} \leftarrow \textbf{X} -\text{CoM}(\textbf{X})$.
   \FOR{($i=1; i < n; i++$)}
   \FOR{($j=1; j < k; j++$)}
   \STATE Compute edge-wise equivariant frames $\mathcal{F}_{ij}$ via Eq. \ref{edge frame}: $$\mathcal{F}_{ij}=\mathbf{EquiFrame}(\textbf{x}_i, \textbf{x}_j).$$
   \STATE Get the mutual 3D structure $\textbf{S}_{i-j}$, perform local scalarization:
   $$t_{ij}=\mathbf{Scalarize}(\textbf{S}_{i-j})$$
   \STATE Calculate the SE(3)-invariant structural coefficients: $A_{ij} = g(t_{ij},d_{ij})$
   \STATE Perform invariant message passing:
   $$m_{ij} = \phi_m( h_i, A_{ij} \odot h_j, d_{ij})$$
   \ENDFOR
    \STATE Update invariant node features:
    $$\mathbf{h}_i = \phi_h (\mathbf{h}_i,\sum_{j \in \mathcal{N}(i)} m_{ij})$$
   \ENDFOR
   \STATE {\bfseries Output: $\text{AvgPooling}\ (h_1,\dots,h_n)$.}
\end{algorithmic}
\end{algorithm}

\section{Related Proofs and Discussions of Section \ref{sec:local} }
\label{appendix:isomor}
In section \ref{sec:local}, we proposed a novel hierarchy of local geometric isomorphisms that further motivates the design of incorporating the mutual 3D substructure's information into equivariant GNNs. Different from our fine-grained local characterization, a cocurrent work GWL~\cite{joshi2022expressive}) proposes to measure geometric isomorphism from local to global by the k-hop partition. 

From another point of view, we essentially demonstrated that encoding mutual 3D substructures expands the capacity of the transformation function class with respect to an equivariant GNN.
\citep{wangcomenet} put forward the \textbf{Completeness} concept for characterizing these transformation functions. However, it mainly concentrates on testing whether a function can discriminate global geometric isomorphism (in the sense of Eq. \ref{global}).

\textbf{Discussion on the Completeness Concept.}
Following our terminology in the preliminary section, \textbf{completeness} of a transformation $f$ can be translated into claiming that $f$ is invariant among 3D graphs if and only if they are \textbf{globally} isomorphic (see definition (\ref{global})). Therefore, it's easy to refine the notion of completeness that adapts to our local version by replacing the global isomorphism to local isomorphism:
$$f(\textbf{X}) = f(\textbf{Y}),$$
if and only if $\textbf{X}$ and $\textbf{Y}$ are local \{-tree, -triangular, -subgraph \} isomorphic.
Then, in terms of function class capacities, the following relation holds:
$$\textbf{Global complete} \subset \textbf{Subgraph complete} \subset \textbf{Triangular complete} \subset \textbf{Tree complete}.$$
Note that our equivalent description of complete transformation reveals the fact that the completeness concept in \cite{wangcomenet} is defined from the global 3D isomorphism point of view. Therefore, we shall claim that the above series of completeness notions belong to the \textbf{structure} completeness. Indeed, the theory developed in section \ref{sec:local} indicates that a GNN which can express \textbf{structure} complete functions may not be sufficient in expressing general tensor potential functions on a 3D graph.

On the other hand, a non-negligible proportion of 3D graph tasks may not be sensitive to the global 3D non-isomorphism. For example, some chemical properties (formulated as a function defined on molecular graphs) are characterized by local substructures~\cite{gong2022examining}. In these scenarios, we are looking for a geometric transformation $f$ that is global non-complete, but (-tree, -triangular, -subgraph) local complete. 

\textbf{Proof of Theorem \ref{thm isomorphism}.} 
\begin{proof}
The first part of the theorem is proved by providing an explicit example. From the first 3D shapes of figure \ref{fig:local_isomorphism}, the difference of the two triangular non-isomorphism shapes is indicated by an invariant function:
$$d(\textbf{x}_p,\textbf{x}_m) = \norm{\textbf{x}_p - \textbf{x}_m}^2.$$
Note that this function cannot be expressed by tree-level features, since there is no edge connecting $\textbf{x}_p$ and $\textbf{x}_m$. However, since the position vectors $\textbf{x}_p$ and $\textbf{x}_m$ are included in the mutual 3D structure (corresponds to edge $e_{iq}$), then they are ready to be scalarized by a local equivariant frame. Then quoting the universal approximation theorem from ~\cite{du2022se}, there exists a corresponding invariant encoder $\phi$ that can approximate the function $d(\textbf{x}_k,\textbf{x}_l)$. Since this function produces different output values for the two tree isometric but triangular non-isometric 3D shapes, we know that $\phi$ is able to distinguish 3D shapes beyond tree isomorphism.

To prove the second part and build up the injectivity condition, we first introduce the multi-set notation $\{\!\!\{\cdot\}\!\!\}$, following [GIN]. A basic equivariant GNN based on our enhanced framework contains at least  two steps: 1. Message passing, which is defined by (\ref{weighted message passing}); 2. Node-wise update:
$$h_i^{t+1} = \textbf{MLP}(m_i^t, h_i^t).$$
For simplicity, we denote the composition of the two steps by $\Psi$.
Then the additional injectivity condition is stated as follows:
\begin{equation} \label{injective}
\Psi (\{\!\!\{h_i^t, A_{ji}h_i^t,  ,h_j^t | j \in \mathcal{N}_i\}\!\!\}, \{\!\!\{A_{ij}h_j^t | j \in \mathcal{N}_i\}\!\!\})    
\end{equation}
is injective, for each layer $t$ and each node $i$. Note that this condition is realizable by adding weighted residue terms similar to \cite{xu2018powerful}. Then, from the above condition, it's obvious that two non-identical collections of $\{\{A_{ij}\}_{e_{ij} \in E}\}$  would yield two different feature vectors. Moreover, from the first part, there exist at least two  distinct local 3D subgraphs with isometric local
tree structure, such that the corresponding geometric weights $\{\{A_{ij}\}_{e_{ij} \in E}\}$ that come out of the encoder $\phi$ are different. 
\end{proof}
\begin{algorithm}[tb]
   \caption{Equivariant Design for LEFTNET.}
   \label{alg:equivariant}
\begin{algorithmic}[1]
   \STATE {\bfseries Input:} Complete 3D gragh with equivariant positions $\mathbf{X}=(\textbf{x}_1,\dots,\textbf{x}_n) \in \mathbb{R}^{n\times 3}$, invariant node features $h_i \in \mathbb{R}^{d}$, invariant relative distances $d_{ij} \in \mathbb{R}^{1}$, \textbf{equivariant} edge features $\textbf{e}_{ij} \in \mathbb{R}^{c}$. 
   \STATE Centralize the positions: $\textbf{X} \leftarrow \textbf{X} -\text{CoM}(\textbf{X})$. 
   \FOR{($i=1; i < n; i++$)}
    \STATE Compute node-wise equivariant frames $\mathcal{F}_{i}$ via Eq.~\ref{eq:node frame} .
   \FOR{($j=1; j < k; j++$)}
   \STATE Compute edge-wise equivariant frames $\mathcal{F}_{ij}$ via Eq. \ref{edge frame}: $$\mathcal{F}_{ij}=\mathbf{EquiFrame}(\textbf{x}_i, \textbf{x}_j)$$ 
   \STATE Get the mutual 3D structure $\textbf{S}_{i-j}$, perform local scalarization through $\mathcal{F}_{ij}$: % \tikzmark{right1} \tikzmark{top1}
   $$t_{ij}=\{\mathbf{Scalarize}(\textbf{S}_{i-j},\mathcal{F}_{ij})\}$$
   \STATE Calculate the SE(3)-invariant structural coefficients: $A_{ij} = g(t_{ij},d_{ij})$ % \tikzmark{right2} \tikzmark{bottom1}
   \STATE Perform equivariant message passing as in Eq.~\ref{weighted message passing}:
   $$\mathbf{m}_{ij} = \phi_m^1( h_i, A_{ij} \odot h_j, d_{ij}) + \phi_m^2( h_i, A_{ij} \odot h_j, d_{ij})\cdot \textbf{e}_{ij} + \phi_m^3( h_i, A_{ij} \odot h_j, d_{ij})\cdot \mathcal{F}_{ij} $$
   \ENDFOR
    \STATE Equivariant message aggregation: $\mathbf{m}_i = \sum_{j \in \mathcal{N}(i)} \mathbf{m}_{ij}$;
    \STATE Transform equivariant node features through $\mathcal{F}_{i}$: % \tikzmark{top2}
    $$t_{i}=\mathbf{Scalarize}(\mathbf{m}_i, \mathcal{F}_{i})$$ 
    \STATE Update invariant node features:
    $$h_i = \phi_h (h_i,t_i)$$
    \STATE {\bfseries Equivariant Output:} Perform tensorization through $\mathcal{F}_{i}$:
    $$\mathbf{h}_i = \mathbf{Tensorize}(h_i, \mathcal{F}_{i}).$$  
   \ENDFOR
   
\end{algorithmic}
\end{algorithm}

\section{Related Proofs and Discussions of Section \ref{sec:build_express} }
\label{appendix:4}

\begin{wrapfigure}[]{r}{0.3\textwidth}
\vspace{-14 pt}
    \centering
    \includegraphics[width=0.3\textwidth]{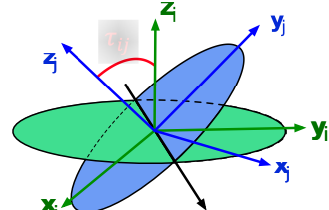}
    \vspace{-20 pt}
    \caption{$\tau_{ij}$ indicates the relative rotation of two frames along the z-axis.
    }\label{fig:euler_angle}
    \vspace{-10 pt}
\end{wrapfigure}
\textbf{Torsion Angle is Secretly Hidden in FT}

Recall the edge-wise (signed) torsion angle $\tau_{ij}$ involves the 1-hop atom pairs $i$ and $j$ and two 2-hop atoms $k$ and $l$, then $\tau_{ij}$ is defined to be the dihedral angle between plane $k-i-j$ and plane $l-j-i$. Although exhausting all torsion angles requires $O(k^2)$ complexity, \cite{wangcomenet} reduces the computation to $O(k)$ order by selecting a canonical 2-hop atom $k$ and $l$, which is enough for detecting the relative orientations between atoms (insufficient for general tasks like many body interactions). 

Now we show how $\tau_{ij}$ naturally appears as one of the derivatives from frame transition functions. For node i, define the equivariant frame $\mathcal{F}_i$ by
$$(\textbf{e}_1^i, \textbf{e}_2^i, \textbf{e}_3^i) = (\textbf{x}_i - \textbf{x}_j, \textbf{x}_i - \textbf{x}_k, \textbf{e}_1^i \times \textbf{e}_2^i).$$
$\mathcal{F}_i$ is normalized through the Gram-Schmidt algorithm. For node j, $\mathcal{F}_j$ is defined similarly by
$$(\textbf{e}_1^j, \textbf{e}_2^j, \textbf{e}_3^j) = (\textbf{x}_j - \textbf{x}_i, \textbf{x}_j - \textbf{x}_l, \textbf{e}_1^j \times \textbf{e}_2^j).$$
Then following the transition formula (\ref{transition}), 
$$R_{ij} = (\textbf{e}_1^i, \textbf{e}_2^i, \textbf{e}_3^i) \cdot (\textbf{e}_1^j, \textbf{e}_2^j, \textbf{e}_3^j)^T.$$
Note that for an orthonormal matrix, its inverse equals its transpose. Then, by the standard definition of a dihedral angle, we have
$$\tau_{ij} = \textbf{e}_3^i \cdot \textbf{e}_3^j \equiv R_{ij}(3,3).$$
In conclusion, $\tau_{ij}$ is just one component of the transition matrix. However, to fully determine $R_{ij}$, we still need another two angles (since a transition matrix is uniquely determined by three Euler angles).

% Indeed, $R_{xy} = U\cdot V$, if we set $U = (e_1^x, e_2^x, e_3^x)$ and $V = (e_1^y, e_2^y, e_3^y)^T$. 

% Since the inner product operation is compatible with our algorithm, we also include it in our node updating block as an additional option.

\textbf{Proof of Theorem \ref{localtoglobal}.} 
\begin{proof}
This theorem is proved in two steps:
\begin{enumerate}
    \item The first step characterizes all scalars determined by (isolated) local clusters $\textbf{B}$ and $\textbf{C}$ through equivariant frames and local scalarization;
    \item The second step constructs a specific invariant function $f_a(\textbf{B},\textbf{C})$ that cannot be expressed by the local scalarization.
\end{enumerate}
Let $G$ denote a 3D point cloud. Then, as it has been proved in \citet{du2022se}, once equipped with an equivariant frame $\mathbf{F}_G$, all equivariant features of $G$ can be transformed to scalar features through scalarization without information loss. Following the convention in the main text, let $\Tilde{G}$ be the output of performing scalarization on the original features $G$, then for any invariant function $f(G)$, there exists a corresponding function $\Tilde{f}$ such that
\begin{equation} \label{information loss}
f(G) = \Bar{f}(\Tilde{G}).
\end{equation}
Since $S_G$ is the collection of all invariant scalars produced by $G$, (\ref{information loss}) implies that $S_G$ is generated by $\Tilde{G}$, which is only a finite subset of $S_G$. To apply the above insight to our current theorem, note that we have two different 3D clouds $\textbf{B}$ and $\textbf{C}$. Therefore, we need to build two local equivariant frames $\mathcal{F}_\textbf{B} = (e^\textbf{B}_1,e^\textbf{B}_2,e^\textbf{B}_3)$ and $\mathcal{F}_\textbf{C} = (e^\textbf{C}_1,e^\textbf{C}_2,e^\textbf{C}_3)$. The crucial point is the frame $\mathcal{F}_\textbf{B}$ itself doesn't depend on $\textbf{C}$, and the scalarization through $\mathcal{F}_\textbf{B}$ is only performed on the \textbf{local} point cloud $\textbf{B}$. Operations like scalarizing equivariant information of $\textbf{C}$ through $\mathcal{F}_\textbf{B}$ would break the assumptions of the theorem. 

Now we are ready to construct an explicit interaction potential $f_a(\textbf{B},\textbf{C})$:
$$f_a(\textbf{B},\textbf{C}) : = e^\textbf{B}_1 \cdot e^\textbf{C}_1.$$
Since $f_a$ is an inner product of two equivariant vectors, it's automatically an invariant function. Then we need to check whether $f_a(\textbf{B},\textbf{C})$ is a function of $f_a(S_\textbf{B},S_\textbf{C})$. Note that the equivariant building block of $f_a$ that relates to $\textbf{B}$ is exactly $e^\textbf{B}_1$. Then, following the above local scalarization principle, we scalarize $e^\textbf{B}_1$ through $\mathcal{F}_\textbf{B}$ and get:
$$e^\textbf{B}_1 \rightarrow \Tilde{e}^C_1 = (1,0,0).$$
Similarly, $e^\textbf{C}_1$ is also transformed to a constant scalar tuple $\Tilde{e}^\textbf{C}_1 = (1,0,0)$ through $\mathcal{F}_\textbf{C}$. As constant inputs generate constant outputs, we conclude that the deduced local scalars can only approximate constant functions. However, since the local frames are changing as we vary the 3D structure of $\textbf{B}$ and $\textbf{C}$, it's obvious that $e^\textbf{B}_1 \cdot e^\textbf{C}_1$ is not a constant function of $(\textbf{B},\textbf{C})$. Therefore, we finish the proof by contradiction. 
\end{proof}

\textbf{Realizing FT by Equivariant Messages:\ }
From the \textbf{FT} definition \ref{transition}, each element of the $3 \times 3$ matrix $R_{ij}$ is calculated by
\begin{equation}
R_{ij}(k,l) = \textbf{e}^i_k \cdot \textbf{e}^j_l.
\end{equation}
Now we show how to reproduce $R_{ij}(k,l)$ through equivariant messages. 
Let the equivariant message $\textbf{m}_i$ be the following:
$$\textbf{m}_i : = (\textbf{e}^i_1, \textbf{e}^i_2, \textbf{e}^i_3) \cdot \left[\begin{array}{ccc:ccc:ccc}
1 & 1 & 1 & 0 & 0 & 0 & 0 & 0 & 0\\
\hdashline
0 & 0 & 0 & 1 & 1 & 1 & 0 & 0 & 0\\
\hdashline
0 & 0 & 0 & 0 & 0 & 0 & 1 & 1 & 1\\ 
\end{array}\right]
$$
It's easy to check that $\textbf{m}_i \in \mathbf{R}^{3 \times 9}$ consists of 9 equivariant vectors (\textbf{multi-channels}). For atom $j$, $\textbf{m}_j$ is defined symmetrically. For each node, we also store the scalar messages, e.g., $\norm{\textbf{e}^i_k}$ for $1 \leq k \leq 3$. Flattening the whole matrix $R_{ij}$ into a $\mathbf{R}^{1 \times 9}$ array, then $R_{ij}$ is obtained by simple summation and taking the vector norm:
$$\norm{\textbf{m}_i + \textbf{m}_j} = \left\{\norm{\textbf{e}^i_k + \textbf{e}^j_l}\right\}_{1 \leq k,l \leq 3},$$
where the norm is taken for each column of $\textbf{m}_i + \textbf{m}_j$, such that $\norm{\textbf{m}_i + \textbf{m}_j} \in \mathbf{R}^{1 \times 9}$. Then,
$$R_{ij} = \frac{1}{2}\left[ \norm{\textbf{e}^i_k + \textbf{e}^j_l}^2 - \norm{\textbf{e}^i_k}^2 - \norm{\textbf{e}^j_l}^2\right].$$
Our illustration also demonstrates the importance of keeping multi-channel tensor messages.

\textbf{Relation with Previous Equivariant Update Methods.}
Following the efficiency principle established in section \ref{sec:build_express}, we don't encode the data of the transition matrices explicitly. Instead, we implement tensor messages to fill in the expressiveness gap. Among the tremendously different designs of equivariant graph neural networks, \citet{schutt2021equivariant} is closely related to our equivariant updating method. By the above argument, the inner product operation for node $i$ (see (9) of \citet{schutt2021equivariant}) $$<\textbf{U} \textbf{v}_i , \textbf{V} \textbf{v}_i>$$ can also be reinterpreted as a realization of the (aggregated) frame transition matrix (\ref{transition}). 

Moreover, since the equivariant vectors $\textbf{U}\textbf{v}_i$ and $\textbf{V}\textbf{v}_i$ are both aggregated vector features that belong to the same node $i$ and the inner product operation between them is performed in the node-wise updating phase, \citet{schutt2021equivariant} actually avoids the 2-hop $O(k^2)$ complexity of computing $R_{xy}$ for all neighborhood node pairs $(\textbf{x},\textbf{y})$ (while able to express the torsion angle implicitly). For our algorithm, we utilize the scalarization and tensorization in the node-wise updating phase. By the universal approximation theorem \ref{thm vector update}, our method can approximate any inner product operations.

\section{Related Proofs and Discussions of Section \ref{sec:leftnet}}
\label{appendix:5}

\textbf{Equivariant Frames and Higher Order Scalarization and Tensorization.}
Given an edge $e_{ij}$ with two atom's positions $(\mathbf{x}_i, \mathbf{x}_j)$, our edge-wise $SE(3)$ equivariant frames $\mathcal{F}_{ij}$ are defined by:
\begin{equation} \label{edge frame}
(\textbf{e}_1,\textbf{e}_2,\textbf{e}_3) = (\frac{\mathbf{x}_i - \mathbf{x}_j}{\norm{\mathbf{x}_i - \mathbf{x}_j}}, \frac{\mathbf{x}_i \times \mathbf{x}_j}{\norm{\mathbf{x}_i \times \mathbf{x}_j}}, \frac{\mathbf{x}_i - \mathbf{x}_j}{\norm{\mathbf{x}_i - \mathbf{x}_j}} \times \frac{\mathbf{x}_i \times \mathbf{x}_j}{\norm{\mathbf{x}_i \times \mathbf{x}_j}}).\end{equation}
To make the frame translation invariant, we follow previous works ~\cite{kohler2020equivariant,hoogeboom2022equivariant}
% [(]
% K ̈ohler et al., 2020; Xu et al., 2022; Hoogeboom et al., 2022,Du]
by limiting the
whole 3D conformers' space to a linear subspace where the center of mass (CoM) of the system (either the whole system or the sub-cluster  where $i$ and $j$ belong to) is zero. On the other hand, building an \textbf{E(3)} frame requires an additional atom's position $\mathbf{x}_k$, which can be selected by K-Nearest Neightbor algorithm. Then if $(\mathbf{x}_i,\mathbf{x}_j,\mathbf{x}_k)$ spans the 3D space, we obtain an $E(3)$ equivariant frame by performing the gram-schmidt orthogonalization. 

Once we have an equivariant frame, every vector is a linear combination of the three orthogonal vectors in the frame. Moreover, the unique combination coefficients are exactly the 'scalarized' coordinates in (\ref{scalarization}). A similar procedure also applies to higher order tensors. Indeed, the vector frame $\mathcal{F}^1$ extends to a tensor frame $\mathcal{F}^r$ of arbitrary order $r > 1$:
\begin{equation} \label{tensor frame}
\mathcal{F}^r : = \{\textbf{e}_{1_1} \otimes \cdots \otimes \textbf{e}_{1_r}\}_{1\leq i_1,\dots, i_r \leq 3}.  
\end{equation}
Since the orthonormal frame $\mathcal{F}^r$ is complete in the sense that it spans the whole tensor space of order $r$, every r-th order tensor admits a unique decomposition:
\begin{equation} \label{higher sca}
\mathbf{T} = \sum_{1\leq i_1,\dots, i_r \leq 3} T^{i_1,\dots, i_r} \textbf{e}_{i_1} \otimes \cdots \otimes \textbf{e}_{i_r}.\end{equation}
It's easy to prove that the collection $\{T^{i_1,\dots, i_r}\}_{1\leq i_1,\dots, i_r \leq 3}$ consists of invariant scalars. We call the process from $\mathbf{T}$ to $\{T^{i_1,\dots, i_r}\}_{1\leq i_1,\dots, i_r \leq 3}\ $ \textbf{scalaraization}. 

\textbf{Tensorization} is the inverse of scalarization, in the sense that it sends scalars $\{T^{i_1,\dots, i_r}\}_{1\leq i_1,\dots, i_r \leq 3}\ $ to tensor $\mathbf{T}$. Under the same frames we use during scalarization, the following diagram demonstrates the pipeline of producing $L$ second-order tensors out of $\{T^{i_1i_2}_j\}_{1\leq i_1,i_2 \leq 3}$:
\begin{equation} \label{higher tensorization}
\{\mathbf{T}_1,\dots,\mathbf{T}_L\} = \underbrace{\left \{\begin{bmatrix} T^{11}_1, & T^{12}_1, & T^{13}_1 \\ T^{21}_1, & T^{22}_1, & T^{23}_1 \\ T^{31}_1, & T^{32}_1, & T^{33}_1 \end{bmatrix},\dots,\begin{bmatrix} T^{11}_L, & T^{12}_L, & T^{13}_L \\ T^{21}_L, & T^{22}_L, & T^{23}_L \\ T^{31}_L, & T^{32}_L, & T^{33}_L \end{bmatrix}\right \}}_{L\text{ channels}}  \odot \begin{bmatrix} \textbf{e}_1 \otimes \textbf{e}_1, & \textbf{e}_1 \otimes \textbf{e}_2, & \textbf{e}_1 \otimes \textbf{e}_3 \\ \textbf{e}_2 \otimes \textbf{e}_1, & \textbf{e}_2 \otimes \textbf{e}_2, & \textbf{e}_2 \otimes \textbf{e}_3 \\ \textbf{e}_3 \otimes \textbf{e}_1, & \textbf{e}_3 \otimes \textbf{e}_2, & \textbf{e}_3 \otimes \textbf{e}_3 \end{bmatrix},\end{equation}
where $\odot$ denotes the element-wise product. 
\paragraph{Proof of Theorem \ref{thm vector update}}
\begin{proof}
The proof is based on the fact that \textbf{Scalarization} and \textbf{Tensorization} are invertible (see Appendix A.5 of \citet{du2022se} ). In other words, we have the following commutative diagram:
\begin{tikzcd} \label{commutative2}
     \textbf{T}^{l-1} \arrow[r,  "\rho"] \arrow[d, "\textbf{Scalarize}"]
      & \textbf{T}^{l}  \\
     \Tilde{T}^{l-1} \arrow[r, "\textbf{MLP}"]
      & \Tilde{T}^{l-1} \arrow[u,  "\textbf{Tensorize}"'].
\end{tikzcd}
Therefore, for each mapping $\rho$, we can always find a corresponding`scalarized' mapping $\Tilde{\rho}$:
$$\Tilde{\rho} : = \textbf{Tensorize} \circ \rho \circ \textbf{Scalarize}.$$
Now we have turned from expressing equivariant $\rho$ to the invariant $\Tilde{\rho}$. Note that \textbf{MLP} is a universal approximator of invariant functions, therefore we can always find a \textbf{MLP} to express $\Tilde{\rho}$. By reserving the arrows, we finish the proof. 
\end{proof}
\paragraph{Proof of Equivariance for LEFTNet}
LEFTNet consists of multiple layers of \textbf{LSE} and \textbf{FTE}. \textbf{LSE} is realized by scalarization, and \textbf{FTE} is realized by scalarization and tensorization. Since the invariance of scalarization and the equivariance of tensorization have been proved, we finish the proof.

\section{Additional Experimental Results}
\label{appendix:exp}

\textbf{Ablation Study.} As discussed in Section~\ref{sec:leftnet}, there are two main modules in LEFTNet, namely \textbf{LSE} and \textbf{FTE}. We conduct experiments on QM9 and MD17 to show the importance of each component. Experimental results are summarized in Table~\ref{tab:ablation_qm9} and Table~\ref{tab:ablation_md17}.
The results show that using \textbf{LSE} can outperform the model without both \textbf{LSE} and \textbf{FTE} on all tasks.
Adding \textbf{FTE} can further improve the performance. 
The results demonstrate the importance of \textbf{LSE} and \textbf{FTE} modules.

\begin{table*}[ht]
  \centering
  \caption{Ablation study on QM9 dataset. The evaluation metric is MAE for each property. The \textbf{best} performances are bolded and the \underline{second best} are underlined.
  }
  \resizebox{0.95\textwidth}{!}{
  \begin{tabular}{l c c c c c c c c c c c c}
  \toprule
    Task & $\alpha$ & $\Delta \varepsilon$ & $\varepsilon_{\mathrm{HOMO}}$ & $\varepsilon_{\mathrm{LUMO}}$ & $\mu$ & $C_{\nu}$ & $G$ & $H$ & $R^2$ & $U$ & $U_0$ & ZPVE \\
    Units & bohr$^3$ & meV & meV & meV & D & cal/mol K & meV & meV & bohr$^3$ & meV & meV & meV \\
    \midrule
    LEFTNet (w/o \textbf{LSE} and \textbf{FTE})  & .053 & 49 & 33 & 25 & .038 & .026 & 9 & 8 & .425 & \underline{8} & 8 & 1.59\\
    LEFTNet (\textbf{LSE} only)                  & .043 & 49 & 31 & 23 & \underline{.031} & \underline{.025} & 8 & 7 & .156 & \underline{8} & 7 & 1.34 \\
    LEFTNet (\textbf{LSE} + vector \textbf{FTE}) & \underline{.039} & \underline{39} & \underline{23} & \underline{18} & \textbf{.011} & \textbf{.022} & \textbf{6} & \textbf{5} & \textbf{.094} & \textbf{5} & \textbf{5} & \textbf{1.19} \\
    LEFTNet (\textbf{LSE} + tensor \textbf{FTE}) & \textbf{.038} & \textbf{38} & \textbf{22} & \textbf{17} & \textbf{.011} & \textbf{.022} & \underline{7} & \underline{6} & \underline{.096} & \textbf{5} & \underline{6} & \underline{1.20}\\
    \bottomrule
  \end{tabular}}
  \label{tab:ablation_qm9}
\end{table*}

\begin{table}[ht]
\centering 
\caption{Abalation Study on MD17 dataset. The evaluation metric is MAE for per-atom forces prediction (kcal/mol \AA). The \textbf{best} performances are bolded and the \underline{second best} are underlined.}
\label{tab:ablation_md17}
\resizebox{0.95\textwidth}{!}{
\begin{tabular}{lcccc}\toprule
% &\multicolumn{4}{c}{WoFE=100} \\\midrule
Molecule        & LEFTNet (w/o \textbf{LSE} and \textbf{FTE}) & LEFTNet (\textbf{LSE} only) & LEFTNet (\textbf{LSE} + vector \textbf{FTE}) & LEFTNet (\textbf{LSE} + tensor \textbf{FTE})  \\\midrule
Aspirin         & 1.083 &  0.451 & \underline{0.300} & \textbf{0.210}  \\
Benzene         & 0.425 & 0.185 & \textbf{0.145} & \underline{0.176}\\
Ethanol         & 0.341 & 0.149 & \underline{0.138} & \textbf{0.118}\\
Malonaldehyde   & 0.594 & 0.276  & \underline{0.209} & \textbf{0.159}\\
Naphthalene     & 0.658 & 0.175  & \underline{0.073} & \textbf{0.063}\\
Salicylic  acid & 0.828 & 0.313  & \underline{0.167} & \textbf{0.141}\\
Toluene         & 0.625 & 0.166  & \underline{0.084} & \textbf{0.070}\\
Uracil          & 0.581 & 0.206  & \textbf{0.116} & \underline{0.117}\\
\bottomrule
\end{tabular}}
\end{table}

\textbf{Results on rMD17.} Following~\cite{batatiamace}, we conduct experiments on rMD17 to compare with recent studies. Results show that our LEFTNet can achieve comparable performance to state-of-the-art methods such as MACE and NequIP, while outperforming other baseline methods like GemNet and PaiNN.

\begin{table*}[ht]
\centering 
\vspace{-10 pt}
\caption{Mean Absolute Error for energy(meV) per-atom forces prediction (meV \AA) on rMD17 dataset. Baseline results are taken from~\citet{batatiamace}. The best results are \textbf{bolded}.
}\label{tab:exp_rmd17}
\resizebox{0.99\textwidth}{!}{
\begin{tabular}{llccccccccccc}\toprule
& &LEFTNet &MACE &Allegro &BOTNet &NequIP &GemNet (T/Q) &ACE &FCHL &GAP &ANI &PaiNN \\\midrule
\multirow{2}{*}{Aspirin} &E &\textbf{2.1} &2.2 &2.3 &2.3 &2.3 &- &6.1 &6.2 &17.7 &16.6 &6.9 \\
&F &\textbf{6.4} &6.6 &7.3 &8.5 &8.2 &9.5 &17.9 &20.9 &44.9 &40.6 &16.1 \\
\multirow{2}{*}{Azobenzene} &E &\textbf{0.7} &1.2 &1.2 &\textbf{0.7} &\textbf{0.7} &- &3.6 &2.8 &8.5 &15.9 &- \\
&F &3.3 &3.0 &\textbf{2.6} &3.3 &2.9 &- &10.9 &10.8 &24.5 &35.4 &- \\
\multirow{2}{*}{Benzene} &E &0.05 &0.4 &0.3 &\textbf{0.03} &0.04 &- &0.04 &0.35 &0.75 &3.3 &- \\
&F &0.3 &0.3 &\textbf{0.2} &0.3 &0.3 &0.5 &0.5 &2.6 &6 &10 &- \\
\multirow{2}{*}{Ethanol} &E &\textbf{0.4} &\textbf{0.4} &\textbf{0.4} &\textbf{0.4} &\textbf{0.4} &- &1.2 &0.9 &3.5 &2.5 &2.7 \\
&F &3.6 &\textbf{2.1} &\textbf{2.1} &3.2 &2.8 &3.6 &7.3 &6.2 &18.1 &13.4 &10 \\
\multirow{2}{*}{Malonaldehyde} &E &0.8 &0.8 &\textbf{0.6} &0.8 &0.8 &- &1.7 &1.5 &4.8 &4.6 &3.9 \\
&F &5.4 &4.1 &\textbf{3.6} &5.8 &5.1 &6.6 &11.1 &10.3 &26.4 &24.5 &13.8 \\
\multirow{2}{*}{Naphthalene} &E &0.8 &0.5 &\textbf{0.2} &\textbf{0.2} &0.9 &- &0.9 &1.2 &3.8 &11.3 &5.1 \\
&F &1.9 &1.6 &\textbf{0.9} &1.8 &1.3 &1.9 &5.1 &6.5 &16.5 &29.2 &3.6 \\
\multirow{2}{*}{Paracetamol} &E &\textbf{1.3} &\textbf{1.3} &1.5 &\textbf{1.3} &1.4 &- &4 &2.9 &8.5 &11.5 &- \\
&F &\textbf{4.7} &4.8 &4.9 &5.8 &5.9 &- &12.7 &12.3 &28.9 &30.4 &- \\
\multirow{2}{*}{Salicylic acid} &E &0.9 &0.9 &0.9 &0.8 &\textbf{0.7} &- &1.8 &1.8 &5.6 &9.2 &4.9 \\
&F &4.1 &3.1 &\textbf{2.9} &4.3 &4 &5.3 &9.3 &9.5 &24.7 &29.7 &9.1 \\
\multirow{2}{*}{Toluene} &E &\textbf{0.3} &0.5 &0.4 &\textbf{0.3} &\textbf{0.3} &- &1.1 &1.7 &4 &7.7 &4.2 \\
&F &2.2 &\textbf{1.5} &1.8 &1.9 &1.6 &2.2 &6.5 &8.8 &17.8 &24.3 &4.4 \\
\multirow{2}{*}{Uracil} &E &\textbf{0.4} &0.5 &0.6 &\textbf{0.4} &\textbf{0.4} &- &1.1 &0.6 &3 &5.1 &4.5 \\
&F &2.8 &2.1 &\textbf{1.8} &3.2 &3.1 &3.8 &6.6 &4.2 &17.6 &21.4 &6.1 \\
\bottomrule
\end{tabular}}
\end{table*}

\textbf{Model and training hyperparameters.} Model and training hyperparameters for our method on different datasets are listed in Table~\ref{tab:hyperpara}.

\begin{table}[ht]\centering
\caption{Model and training hyperparameters for our method on different tasks.}\label{tab:hyperpara}
\resizebox{0.6\textwidth}{!}{
\begin{tabular}{lcccc}\toprule
\multirow{2}{*}{Hyperparameter} &\multicolumn{4}{c}{Values/Search Space} \\
\cmidrule{2-4}
&QM9 &MD17 &rMD17 \\
\midrule
Number of layers & 4, 5, 6 &4, 6 &4, 6 \\
Hidden channels & 128, 192, 256 &256 &256 \\
Number of radial basis & 24, 32, 96 &16, 32, 64 &16, 32, 64 \\
Cutoff & 5, 6, 6.5, 8 &6, 8, 10 &6, 8, 10 \\
Epochs & 800 &1000 &1000 \\
Batch size & 32 &1, 4 &1, 4 \\
Learning rate & 1e-4, 5e-4 &5e-4 &5e-4 \\
Learning rate scheduler & steplr &steplr &steplr \\
Learning rate decay factor & 0.5 &0.5 &0.5 \\
Learning rate decay epochs & 100 &200 &200 \\
\bottomrule
\end{tabular}}
\end{table}

\end{document}